\documentclass{article} 
\usepackage[final]{colm2025_conference}

\usepackage{microtype}
\usepackage{hyperref}
\usepackage{url}
\usepackage{booktabs}

\usepackage{lineno}
\usepackage{hyperref}
\usepackage{url}
\usepackage{fancyhdr}
\usepackage{tikz}
\usepackage{hyperref}
\usepackage{url}
\usepackage{algorithm}
\usepackage{algorithmic}
\usepackage{enumitem}
\definecolor{light_gray}{HTML}{FFFFFF} 
\definecolor{solid_gray}{HTML}{000000} 
\usepackage{multirow}
\usepackage{array}
\usepackage{hhline}
\usepackage{pifont}
\usepackage{makecell}
\usepackage{array} 
\usepackage{booktabs}
\usepackage{siunitx}
\usepackage{nameref}
\usepackage{xcolor}
\usepackage{transparent}
\usepackage{soul}
\usetikzlibrary{calc}
\usepackage{tcolorbox}
\usepackage{amssymb}
\usepackage{bbding}
\usepackage{pifont}
\usepackage{longtable}
\usepackage{tabularx}
\usepackage{adjustbox}
\usepackage{caption}
\usepackage{adjustbox}
\usepackage{pgfplots}
\usepackage{subcaption}
\usepackage{caption}
\usepackage{xcolor}
\usepackage{lipsum}
\usepackage{wrapfig}
\usepackage{pgfplots}
\usetikzlibrary{positioning, arrows.meta}

\definecolor{color1}{HTML}{000000} 
\definecolor{color2}{HTML}{E41A1C} 
\definecolor{color3}{HTML}{009E73} 
\definecolor{color4}{HTML}{0072B2} 

\definecolor{color5}{HTML}{555555} 

\definecolor{color6}{HTML}{FF8C00} 
\definecolor{color7}{HTML}{E75480} 

\definecolor{transblue}{rgb}{0.69, 0.85, 0.96} 
\definecolor{editcolor}{rgb}{0.7, 0, 0} 

\newtcbox{\GreenHighlight}[1][]{%
    on line,
    arc=0pt,
    outer arc=0pt,
    colback=green!15,
    boxsep=0pt,
    left=2pt,
    right=2pt,
    top=2pt,
    bottom=2pt,
    boxrule=0pt,
    before=\strut, 
    after=\strut, 
    #1
}

\newtcbox{\BlueHighlight}[1][]{%
    on line,
    arc=0pt,
    outer arc=0pt,
    colback=transblue!30,
    boxsep=0pt,
    left=2pt,
    right=2pt,
    top=2pt,
    bottom=2pt,
    boxrule=0pt,
    before=\strut, 
    after=\strut, 
    #1
}

\newtcbox{\PinkHighlight}[1][]{%
    on line,
    arc=0pt,
    outer arc=0pt,
    colback=color7!15,
    boxsep=0pt,
    left=2pt,
    right=2pt,
    top=2pt,
    bottom=2pt,
    boxrule=0pt,
    before=\strut, 
    after=\strut, 
    #1
}

\newtcbox{\GrayHighlight}[1][]{%
    on line,
    arc=0pt,
    outer arc=0pt,
    colback=color5!15,
    boxsep=0pt,
    left=2pt,
    right=2pt,
    top=2pt,
    bottom=2pt,
    boxrule=0pt,
    before=\strut, 
    after=\strut, 
    #1
}

\definecolor{darkblue}{rgb}{0, 0, 0.5}
\hypersetup{colorlinks=true, citecolor=darkblue, linkcolor=darkblue, urlcolor=darkblue}

\title{Memorization in In-Context Learning}

\author{Shahriar Golchin\textsuperscript{\textnormal{1}},
        Mihai Surdeanu\textsuperscript{\textnormal{1}},
        Steven Bethard\textsuperscript{\textnormal{2}},
        Eduardo Blanco\textsuperscript{\textnormal{1}},
        Ellen Riloff\textsuperscript{\textnormal{1}} \\
        \textsuperscript{\textnormal{1}}Department of Computer Science, University of Arizona \\
        \textsuperscript{\textnormal{2}}School of Information, University of Arizona}

\begin{document}

\ifcolmsubmission
\linenumbers
\fi

\maketitle

\pagestyle{fancy}
\fancyhead{} 
\fancyhead[L]{Preprint} 

\begin{abstract}
In-context learning (ICL) has proven to be an effective strategy for improving the performance of large language models (LLMs) with no additional training. However, the exact mechanism behind this performance improvement remains unclear. This study is the first to show how ICL surfaces memorized training data and to explore the correlation between this memorization and performance on downstream tasks across various ICL regimes: zero-shot, few-shot, and many-shot. Our most notable findings include: (1) ICL significantly surfaces memorization compared to zero-shot learning in most cases; (2) demonstrations, without their labels, are the most effective element in surfacing memorization; (3) ICL improves performance when the surfaced memorization in few-shot regimes reaches a high level (about 40\%); and (4) there is a very strong correlation between performance and memorization in ICL when it outperforms zero-shot learning. Overall, our study uncovers memorization as a new factor impacting ICL, raising an important question: to what extent do LLMs truly generalize from demonstrations in ICL, and how much of their success is due to memorization?
\end{abstract}

\section{Introduction}

In-context learning (ICL) has emerged as a powerful method for improving the performance of large language models (LLMs) without extra training \citep{brown2020language}. This method involves including a few task-specific examples, known as demonstrations or shots, within the input prompt, which enables the LLM to infer the target task and generate improved responses. With long-context LLMs \citep[inter alia]{DBLP:journals/corr/abs-2303-08774,DBLP:journals/corr/abs-2312-11805,Lu2023togethercomputer}, ICL has evolved to incorporate hundreds or even thousands of demonstrations, leading to greater performance improvements \citep{DBLP:journals/corr/abs-2405-00200,DBLP:journals/corr/abs-2404-11018,DBLP:journals/corr/abs-2305-15005}.
However, despite its widespread use and straightforward nature, the underlying principles of ICL and its performance improvement capabilities remain unclear \citep[inter alia]{DBLP:conf/emnlp/MinLHALHZ22, DBLP:conf/icml/OswaldNRSMZV23,DBLP:conf/emnlp/RazeghiL0022}.

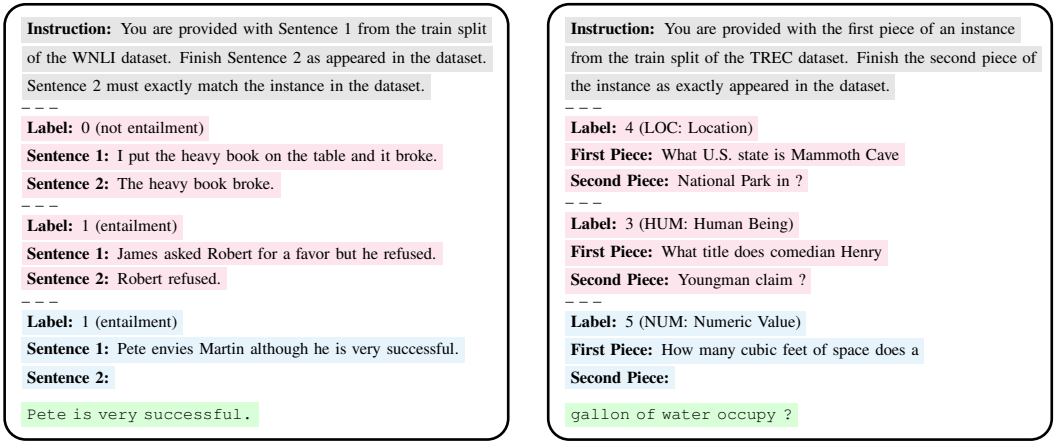
\begin{figure}[!]
\hspace*{0.99\fill}
    \begin{minipage}{0.48\textwidth}
        \centering
        \begin{tikzpicture}[rounded corners=8pt, thick, text=black, text opacity=1]
            \node[draw=solid_gray, fill=light_gray, line width=1pt, text=black, text width=0.95\textwidth, align=left, font=\fontsize{6.2pt}{1em}\linespread{0.8}\selectfont, inner xsep=6.5pt, inner ysep=5pt] at (0,0)
            {
            \GrayHighlight{\textbf{Instruction:}
            You are provided with Sentence 1 from the train}
            \GrayHighlight{split of WNLI dataset. Finish Sentence 2 as appeared in dataset.}
            \GrayHighlight{Sentence 2 must exactly match the instance in the dataset.}

            -- -- --

            \PinkHighlight{\textbf{Label:} 0 (not entailment)}
            
            \PinkHighlight{\textbf{Sentence 1:} I put the heavy book on the table and it broke.}
            
            \PinkHighlight{\textbf{Sentence 2:} The heavy book broke.}

            -- -- --

            \PinkHighlight{\textbf{Label:} 1 (entailment)}
            
            \PinkHighlight{\textbf{Sentence 1:} James asked Robert for a favor but he refused.}
            
            \PinkHighlight{\textbf{Sentence 2:} Robert refused.}

            -- -- --

            \BlueHighlight{\textbf{Label:} 1 (entailment)}
            
            \BlueHighlight{\textbf{Sentence 1:}
            Pete envies Martin although he is very successful.}
            
            \BlueHighlight{\textbf{Sentence 2:}}

            \vspace{4pt}
            
            \GreenHighlight{\texttt{Pete is very successful.}}
            };
        \end{tikzpicture}
    \end{minipage}\hfill
    \begin{minipage}{0.48\textwidth}
        \centering
        \begin{tikzpicture}[rounded corners=8pt, thick, text=black, text opacity=1]
            \node[draw=solid_gray, fill=light_gray, line width=1pt, text=black, text width=0.95\textwidth, align=left, font=\fontsize{6.2pt}{1em}\linespread{0.8}\selectfont, inner xsep=6.5pt, inner ysep=5pt] at (0,0) 
            {
            \GrayHighlight{\textbf{Instruction:}
            You are provided with the first piece of an }
            \GrayHighlight{instance from the train split of the TREC dataset. Finish the}
            \GrayHighlight{second piece of the instance as exactly appeared in the dataset.}

            -- -- --

            \PinkHighlight{\textbf{Label:} 4 (LOC: Location)}

            \PinkHighlight{\textbf{First Piece:}
            What U.S. state is Mammoth Cave}
            
            \PinkHighlight{\textbf{Second Piece:}
            National Park in ?}

            -- -- --

            \PinkHighlight{\textbf{Label:} 3 (HUM: Human Being)}

            \PinkHighlight{\textbf{First Piece:}
            What title does comedian Henry}

            \PinkHighlight{\textbf{Second Piece:}
            Youngman claim ?}

            -- -- --

            \BlueHighlight{\textbf{Label:} 5 (NUM: Numeric Value)}
            
            \BlueHighlight{\textbf{First Piece:}
            How many cubic feet of space does a}

            \BlueHighlight{\textbf{Second Piece:}}

            \vspace{4pt}
            
            \GreenHighlight{\texttt{gallon of water occupy ?}}
            
            };
        \end{tikzpicture}
    \end{minipage}
    \hspace*{0.99\fill} 
    \caption{\textbf{Illustrative examples of a two-shot ICL prompt for replicating instances from NLI (left) and classification (right) tasks.} Note that, in our actual experiments, we use $k$-shot ICL, where $k = \{0, 25, 50, 100, 200\}$. All colored segments, except the green one, form the input prompt. Specifically, the gray segments indicate the instruction, the red segments display the two demonstrations, the blue segments correspond to the dataset instance being replicated, and the green segment exhibits the generated completion by the underlying LLM (GPT-4) for the subsequent segment of the dataset instance being replicated. For both examples, the generated completions are exact matches.}
    \label{fig:input-prompt-examples}
\end{figure}

In this work, we further study the inner workings of ICL by investigating the previously unexplored relationship between \textit{ICL} and \textit{memorization} of training data in LLMs, and how this memorization correlates with performance. In particular, to show how ICL surfaces memorization, we replace the \emph{learning component} (target variable) in ICL with a \textit{text completion task} which is based solely on \emph{memorization}. To achieve this, we adapt the data contamination detection method proposed by \citet{DBLP:journals/corr/abs-2308-08493}. This method aims to replicate dataset instances through \textit{memorization} to verify their presence in the training data. The process begins by splitting a dataset instance into two random-length segments. The initial segment and the corresponding label of the dataset instance are integrated into the input prompt, instructing the LLM to generate the subsequent segment. The generated completion is then evaluated against the original subsequent segment and categorized as an exact, near-exact, or inexact match, with the first two indicating memorization. To implement this for ICL, we use the same strategy to replicate dataset instances, but with a tweak: we include a few pairs of initial and subsequent segments from different dataset instances, along with their labels in the input prompt, as \textit{demonstrations}. Specifically, each demonstration consists of (1) a pair of initial and subsequent segments, and (2) a label. We then quantify the memorization across various regimes (i.e., zero-shot, few-shot, and many-shot) by counting the number of exact and near-exact matches. Figure \ref{fig:input-prompt-examples} shows prompts for an illustrative two-shot ICL to replicate a dataset instance.


We examine memorization across various $k$-shot scenarios, where $k = \{0, 25, 50, 100, 200\}$.
Here, demonstrations for smaller $k$ values are subsets of those used for larger $k$ values.
We categorize our experiments into three \textit{regimes} based on the values of $k$: zero-shot ($k = 0$), few-shot ($k = \{25, 50\}$), and many-shot ($k = \{100, 200\}$).
Each regime is analyzed under different \textit{settings} to identify key element contributing the most to memorization in ICL.
These elements include (1) instruction, (2) segment pairs, and (3) their respective labels, with the latter two forming the demonstrations.
We vary the amount of in-context information in each setting by selectively including or excluding these elements in the prompt to identify the impact of each element on memorization. First, all elements are included---an instruction containing dataset-specific information (i.e., dataset and partition name) and segment pairs with their labels---to establish an upper bound for memorization level. Figure \ref{fig:input-prompt-examples} depicts this setting. 
Second, we remove the instruction (gray parts in Figure \ref{fig:input-prompt-examples}). Third, we exclude the instruction and labels, leaving only segment pairs (red parts in Figure \ref{fig:input-prompt-examples}, without labels).
Finally, we examine the correlation between the surfaced memorization and performance across all our settings.


The primary contributions of this paper are as follows:
\begin{enumerate}
    \item For the first time, we study the relationship between ICL and memorization.
    \item Our study identifies key element contributing to surfacing memorization in ICL.
    \item We explore the correlation between memorization and performance in ICL. 
    \item By analyzing the surfaced memorization levels in ICL, we identify cases where ICL either succeeds or fails to outperform zero-shot learning.
\end{enumerate}


We made several important observations, summarized in the key findings below:\footnote{See Section \ref{sec:results-and-discussion} for a comprehensive list of observations.}

\begin{enumerate}
    \item ICL with only a few demonstrations (e.g., 25 shots) surfaces significant memorization in most cases for data that is part of the training set.
    \item \textit{Segment pairs---demonstrations without their labels---}are the key element contributing to surfacing memorization in ICL.
    \item \textit{There is a very strong correlation between performance on downstream tasks and surfaced memorization by ICL} when it improves performance compared to zero-shot learning.
    \item ICL outperforms zero-shot learning when the surfaced memorization level in few-shot regimes is significant, particularly at levels around 40\% or more. 
    \item As demonstrations increase, even though the surfaced memorization by ICL remains relatively constant at a high level in most many-shot regimes, near-exact matches gradually become exact matches, making memorization more explicit.
    \item Evaluating performance on memorized and non-memorized instances in ICL reveals that the performance on memorized instances is consistently higher than non-memorized instances across nearly all regimes, from zero-shot to many-shot.
    \item Consistent with the findings of \citet{DBLP:conf/iclr/CarliniIJLTZ23} on memorization in language models, we discovered that memorization significantly increases with the number of tokens of context used to prompt the model. However, our experiments further this finding by showing that these \textit{tokens can be from individual instances}, not only tokens from a single instance.
\end{enumerate}

\section{Terminology}
\label{sec:terminology}

Before presenting our methodology, we establish specific terminology for clarity.

\textbf{Element:} We use the term ``element'' to refer to any of the following: instruction, segment pairs, or labels. Our experiments evaluate how each element affects memorization in ICL.

\textbf{Setting:} One of the key objectives of this study is to identify the main element influencing memorization in ICL. For this, we experiment with three settings, each varying by the amount of in-context information in the input prompt. Therefore, the term ``setting'' refers to \textit{the amount of information incorporated into the input prompt} in our experiments. We detail our settings in Subsection \ref{subsec:identifying-key-component-in-memorization}.


\textbf{Regime:} Contrary to settings, we define regimes based on the values of $k$ in $k$-shot scenarios. Hence, the term ``regime'' emphasizes the \textit{number of demonstrations (shots) used in the input prompt}. We elaborate on these regimes in Subsection \ref{subsec:selection-of-in-context-learning-regimes}.


\textbf{Demonstration:} In the scope of ICL, several terms describe task-specific examples included in the input prompt. For clarity, we use the terms ``demonstrations'' and ``shots'' interchangeably to refer to these examples. As previously noted, each demonstration in our experiments comprises (1) a segment pair with an initial and subsequent segment, and (2) a label. \textit{Therefore, when we mention demonstrations without labels, we only refer to segment pairs.}

\section{Approach}

\subsection{Detecting and Quantifying Memorization}
\label{subsec:detecting-and-quantifying-memorization}

To detect and quantify memorization in ICL, we use the method proposed by \citet{DBLP:journals/corr/abs-2308-08493}, originally designed to detect data contamination in LLMs.
Below, we explain how we adjust the original method to detect memorization in ICL and detail the procedure for quantifying it.

\textbf{Detecting Memorization in In-Context Learning.}
We specifically employ the ``guided instruction'' strategy from \citet{DBLP:journals/corr/abs-2308-08493}. This approach aims to verify if specific instances from a particular dataset partition (e.g., test set) were included in the model's training data by replicating them through memorization. To this end, each dataset instance is split into two random-length segments, and the LLM is then tasked with completing the subsequent segment based on the initial segment and the respective label provided in the input prompt. The prompt also incorporates dataset-specific details (i.e., dataset and partition name) to better guide the LLM in the replication process.

To adapt this strategy for $k$-shot ICL, we include $k$ pairs of initial and subsequent segments from $k$ distinct dataset instances, along with their labels, as \textit{demonstrations} in the input prompt. With this prompt, we follow the same process of replicating the subsequent segments for the dataset instances under consideration. Finally, as in \citet{DBLP:journals/corr/abs-2308-08493}, the similarity between the generated completions and the original subsequent segments is evaluated to determine if the dataset instances were part of the model's training data. Figure \ref{fig:input-prompt-examples} provides examples of demonstrations and their integration into our replication process to study memorization in ICL.


\textbf{Evaluating Memorization in In-Context Learning.} 
\citet{DBLP:journals/corr/abs-2308-08493} proposed three categories for evaluating generated completions to the original subsequent segments:

\noindent \textbf{(1) Exact Match:} The completion exactly matches the original subsequent segment.

\noindent \textbf{(2) Near-Exact Match:} The completion, while not identical, shows considerable overlap and maintains significant semantic and structural similarity to the original subsequent segment.\footnote{Examples of exact and near-exact matches are shown in Table \ref{tab:turning-near-exact-atches-to-exact-matches} in Appendix \ref{app:exact-and-near-exact-matches}.}

\noindent \textbf{(3) Inexact Match:} The completion differs entirely from the original subsequent segment.

They employed GPT-4 with few-shot ICL to classify generated completions into these categories. In particular, this classifier uses a few human-annotated examples of exact and near-exact matches in the prompt as references and automatically compares the generated completions to their original counterparts.\footnote{Figure \ref{fig:few-shot-in-context-evaluation-prompt} in Appendix \ref{appendix:few-shot-in-context-learning-evaluation-prompt} illustrates this evaluation prompt.} We adopt the same method and adhere to the same categories for our evaluation. Although their results showed this evaluation strategy achieves high accuracy (92\%--100\%) in matching evaluations from human judgments, we conduct an additional human evaluation on top of GPT-4's evaluation to ensure optimal accuracy in our findings. This is important as our conclusions significantly rely on the number of detected exact and near-exact matches. We detail our human evaluation process in Section \ref{sec:experimental-setup}, under Human Evaluation.



\textbf{Quantifying Memorization in In-Context Learning.} Following \citet{DBLP:journals/corr/abs-2308-08493}, we consider both exact and near-exact matches as indicators of memorization. We quantify memorization by counting the number of these matches and expressing them as a percentage of the total dataset instances under consideration. 

\subsection{Identifying Key Element in Memorization in In-Context Learning}
\label{subsec:identifying-key-component-in-memorization}

Our experiments involve three distinct settings, all aiming at quantifying memorization but differing in the amount of information included in the input prompt. This helps us measure the amount of memorization in ICL regimes based on the information provided by each element and identify the key element in the process. As shown in Figure \ref{fig:input-prompt-examples} and discussed in Section \ref{sec:terminology}, the input prompt is composed of two main parts: the \textit{instruction}, which contains dataset-specific details, and \textit{demonstrations}, which include \textit{segment pairs} and their respective \textit{labels}. We combine these three elements---\textit{instruction}, \textit{segment pairs}, and \textit{labels}---in different ways to create three unique settings with varying amounts of in-context information. Below, we detail each setting.

\textbf{(1) Full Information.} This setting maximizes in-context information by \textit{including all three elements: instruction, segment pairs, and labels}. Figure \ref{fig:input-prompt-examples} illustrates this setting. In fact, this setting contains more information than standard ICL by incorporating dataset-specific details not typically included. We use it to establish an upper bound for the highest possible amount of memorization that can be surfaced in ICL regimes. By comparing the impact of each element on memorization against this maximum, we identify which element most significantly influences memorization in ICL.

\textbf{(2) Segment Pairs and Labels.} Here, we exclude the instruction containing dataset-specific information and \textit{include only segment pairs and labels}. To show this setting, it omits the gray segments in Figure \ref{fig:input-prompt-examples} and includes only the red segments. This setting is closest to standard ICL, although standard ICL includes an instruction for executing the target task, which is absent here. However, since this instruction lacks relevant information that can affect memorization, its impact on memorization is zero or negligible. Additionally, as we see in Section \ref{sec:results-and-discussion}, even an instruction with dataset-specific information (as in the previous setting) has minimal impact on memorization in ICL regimes.

\textbf{(3) Only Segment Pairs.} We further remove elements from the input prompt and \textit{include only segment pairs}, excluding the instruction and labels. While the previous setting examines the combined effect of segment pairs and labels on memorization in ICL, this setting shows their individual contributions. By comparing the amount of surfaced memorization in this setting with the one that includes both segment pairs and labels, as well as the full information setting, we can assess how much memorization is due to segment pairs alone versus labels. This helps identify the primary element driving memorization in ICL regimes.


\subsection{Performance and Memorization in In-Context Learning}
As the primary goal of using ICL is to enhance downstream performance, we explore the connection between memorization in ICL and performance. We compute the performance on the samples for which we assess memorization and analyze the correlation between performance and memorization across our three settings using the Pearson correlation \citep{pearson1895vii}. In addition, we separately evaluate performance for \textit{memorized} and \textit{non-memorized} instances across ICL regimes to further explore this relationship. According to Subsection \ref{subsec:detecting-and-quantifying-memorization}, instances that are replicated exactly or nearly exactly are considered memorized, while those replicated inexactly are considered non-memorized. Note that, for performance measurement, we use standard $k$-shot ICL, which includes an instruction to perform the task with $k$ demonstrations and their labels embedded in the input prompt.

\subsection{Selection of In-Context Learning Regimes}
\label{subsec:selection-of-in-context-learning-regimes}

We work with five $k$-shot ICL regimes across our three settings, where $k = \{0, 25, 50, 100, 200\}$, covering zero-shot, few-shot, and many-shot regimes. Specifically, we define zero-shot regimes when $k = 0$, few-shot regimes when $k = \{25, 50\}$, and many-shot regimes when $k = \{100, 200\}$.  In our experiments, to assess the impact of increasing demonstrations on memorization and performance, we progressively increase the number of demonstrations, ensuring that larger regimes include all demonstrations from the smaller ones. For example, the 100-shot ICL includes 50 demonstrations from the 50-shot ICL, which itself includes 25 demonstrations from the 25-shot ICL. Details on how demonstrations are prepared and used are provided in the Experimental Setup section in Appendix~\ref{sec:experimental-setup}.

\subsection{Selection of Models}
\label{subsec:selection-of-models}

To achieve the goals of our study, LLMs must meet specific criteria to be selected. First, they must be highly performant, with strong steerability and controlled generation, allowing us to systematically quantify memorization through their outputs.
This is crucial given the opaque nature of the training data---if a model fails to replicate a dataset instance, we can reasonably conclude it was not part of the training data, rather than attributing it to model's inability to replicate.
Less performant models may keep memorization internal by not explicitly emitting memorized data, or generate outputs that are too unstructured to detect memorization effectively.
Second, as we extend our experiments to many-shot regimes, LLMs must support long contexts to accommodate our largest regime with 200 demonstrations across all datasets. Third, candidate LLMs must have been trained on diverse datasets. This diversity is key for observing how memorization evolves across different ICL regimes via instance replication. Clearly, without this, studying memorization is unfeasible.
Note that, if an LLM does not meet these criteria, it does not invalidate our conclusions. In fact, these criteria are essential for effectively \textit{studying} memorization, but memorization exists in all language models regardless \citep{DBLP:conf/iclr/CarliniIJLTZ23,DBLP:conf/uss/CarliniTWJHLRBS21}. Additional information on our pilot study and the selection of GPT-4 for our experiments is provided in the Experimental Setup section in Appendix~\ref{sec:experimental-setup}.

\subsection{Selection of Datasets}
\label{subsec:selection-of-datasets}

Following the settings described in Subsection \ref{subsec:identifying-key-component-in-memorization}, datasets in our study must fulfill certain criteria.
First, datasets must be part of the training corpora for the LLMs used in our study, ensuring that their instances can be potentially replicated via memorization.
Second, to evaluate the impact of labels on memorization in ICL regimes, we need datasets with labeled samples. Third, these datasets should have a complex label space or be challenging enough for LLMs, allowing us to observe performance change across ICL regimes and explore its correlation with memorization. Fourth, the sample length must be limited to a few dozen tokens to fit within the input context length of LLMs for all datasets, handling up to 200 demonstrations in our largest many-shot regime. Further details on our pilot study and the dataset selection process can be found in the Experimental Setup section in Appendix~\ref{sec:experimental-setup}.


\section{Results and Discussion}
\label{sec:results-and-discussion}

We first discuss our results on memorization across various ICL regimes in our three settings. We then examine how this memorization correlates with performance on downstream tasks.\footnote{In Appendix \ref{sec:comparing-observations-with-previous-studies}, we compare our observations with prior studies showcasing specific characteristics of ICL. Additionally, we discuss the practical implications of our observations in Appendix \ref{app:practical-implications}.}

\subsection{Results on Quantifying Memorization}
\label{subsec:quantifying-memorization}

\begin{wrapfigure}[39]{r}{0.66\textwidth}
    \centering
    \vspace{-\baselineskip}

    
    \begin{center}
        \parbox{0.5\textwidth}{\centering \small\textbf{(1) Full Information}}
    \end{center}
    \vspace{-0.4cm}
    
    \begin{subfigure}[b]{0.48\linewidth}
        \centering
        \begin{tikzpicture}
            \begin{axis}[
                width=\linewidth,            
                height=\linewidth,           
                xlabel={\scriptsize Number of Shots ($k$-shot)},  
                ylabel={\tiny Exact \& Near-Exact Matches (\%)},  
                ylabel style={yshift=-1.4em},
                xlabel style={yshift=0.4em},
                xmin=-15, xmax=215,
                ymin=-5, ymax=85,
                ytick={0,10,20,30,40,50,60,70,80},
                legend pos=south east,
                legend cell align={left},
                grid=major,
                xtick={0,25,50,100,200},
                xticklabels={0,25,50,100,200},
                extra x tick labels={$\vert\vert$},
                extra x tick style={grid=none},
                tick label style={font=\scriptsize},
                x tick label style={font=\scriptsize},
                axis line style={line width=0.65pt},
                legend style={
                    font=\scriptsize,
                    nodes={scale=0.45, transform shape},
                    fill=white, fill opacity=0.3, text opacity=1,
                    draw opacity=0.6, draw=white!50!black,
                    rounded corners=2pt
                },
                legend image post style={opacity=1},
            ]
            \addplot[
                densely dashed,
                mark=*,
                color=color1,
                very thick,
                mark options={scale=0.8, solid},
            ]
            coordinates {
                (0,33) (25,59) (50,64) (100,67.5) (200,75)
            };
            \addlegendentry{WNLI}

            \addplot[
                densely dotted,
                mark=square*,
                color=color2,
                very thick,
                mark options={scale=0.73, solid},
            ]
            coordinates {
                (0,31.5) (25,39.5) (50,39) (100,41) (200,41)
            };
            \addlegendentry{TREC}

            \addplot[
                dash pattern=on 6pt off 2pt on 2pt off 2pt,
                mark=triangle*,
                color=color3,
                very thick,
                mark options={scale=1.0, solid},
            ]
            coordinates {
                (0,11) (25,18.5) (50,21) (100,25) (200,25.5)
            };
            \addlegendentry{RTE}

            \addplot[
                densely dashdotted,
                mark=diamond*,
                color=color4,
                very thick,
                mark options={scale=1.0, solid},
            ]
            coordinates {
                (0,47.5) (25,51) (50,50) (100,51.5) (200,53)
            };
            \addlegendentry{DBpedia}
            \end{axis}
        \end{tikzpicture}
    \end{subfigure}
    \begin{subfigure}[b]{0.48\linewidth}
        \centering
        \vspace{1.0em} 
        \begin{tikzpicture}
            \begin{axis}[
                width=\linewidth,
                height=\linewidth,
                xlabel={\scriptsize Number of Shots ($k$-shot)},  
                ylabel={\scriptsize Exact Matches (\%)},          
                ylabel style={yshift=-1.4em},
                xlabel style={yshift=0.4em},
                xmin=-15, xmax=215,
                ymin=-5, ymax=75,
                ytick={0,10,20,30,40,50,60,70},
                legend pos=north west,
                legend cell align={left},
                grid=major,
                xtick={0,25,50,100,200},
                xticklabels={0,25,50,100,200},
                extra x tick labels={$\vert\vert$},
                extra x tick style={grid=none},
                tick label style={font=\scriptsize},
                x tick label style={font=\scriptsize},
                axis line style={line width=0.65pt},
                legend style={
                    font=\scriptsize,
                    nodes={scale=0.45, transform shape},
                    fill=white, fill opacity=0.3, text opacity=1,
                    draw opacity=0.6, draw=white!50!black,
                    rounded corners=2pt
                },
                legend image post style={opacity=1},
            ]
            \addplot[
                densely dashed,
                mark=*,
                color=color1,
                very thick,
                mark options={scale=0.8, solid},
            ]
            coordinates {
                (0,21) (25,45) (50,49.5) (100,55.5) (200,64)
            };
            \addlegendentry{WNLI}

            \addplot[
                densely dotted,
                mark=square*,
                color=color2,
                very thick,
                mark options={scale=0.73, solid},
            ]
            coordinates {
                (0,22.5) (25,31) (50,30) (100,32) (200,32)
            };
            \addlegendentry{TREC}

            \addplot[
                dash pattern=on 6pt off 2pt on 2pt off 2pt,
                mark=triangle*,
                color=color3,
                very thick,
                mark options={scale=1.0, solid},
            ]
            coordinates {
                (0,2.5) (25,6) (50,6.5) (100,6.5) (200,9.5)
            };
            \addlegendentry{RTE}

            \addplot[
                densely dashdotted,
                mark=diamond*,
                color=color4,
                very thick,
                mark options={scale=1.0, solid},
            ]
            coordinates {
                (0,8.5) (25,22.5) (50,24) (100,24.5) (200,24)
            };
            \addlegendentry{DBpedia}
            \end{axis}
        \end{tikzpicture}
    \end{subfigure}
    
    \vspace{-0.2cm}
    
    \begin{center}
        \parbox{0.5\textwidth}{\centering \small\textbf{(2) Segment Pairs and Labels}}
    \end{center}
    \vspace{-1.1cm}
    
    \begin{subfigure}[b]{0.48\linewidth}
        \centering
        \begin{tikzpicture}
            \begin{axis}[
                width=\linewidth,
                height=\linewidth,
                xlabel={\scriptsize Number of Shots ($k$-shot)},
                ylabel={\tiny Exact \& Near-Exact Matches (\%)},
                ylabel style={yshift=-1.4em},
                xlabel style={yshift=0.4em},
                xmin=-15, xmax=215,
                ymin=-5, ymax=85,
                ytick={0,10,20,30,40,50,60,70,80},
                legend pos=south east,
                legend cell align={left},
                grid=major,
                xtick={0,25,50,100,200},
                xticklabels={0,25,50,100,200},
                extra x tick labels={$\vert\vert$},
                extra x tick style={grid=none},
                tick label style={font=\scriptsize},
                x tick label style={font=\scriptsize},
                axis line style={line width=0.65pt},
                legend style={
                    font=\scriptsize,
                    nodes={scale=0.45, transform shape},
                    fill=white, fill opacity=0.3, text opacity=1,
                    draw opacity=0.6, draw=white!50!black,
                    rounded corners=2pt
                },
                legend image post style={opacity=1},
            ]
            \addplot[
                densely dashed,
                mark=*,
                color=color1,
                very thick,
                mark options={scale=0.8, solid},
            ]
            coordinates {
                (0,11) (25,57.5) (50,63.5) (100,66) (200,75)
            };
            \addlegendentry{WNLI}

            \addplot[
                densely dotted,
                mark=square*,
                color=color2,
                very thick,
                mark options={scale=0.73, solid},
            ]
            coordinates {
                (0,16.5) (25,40.5) (50,39) (100,40) (200,40)
            };
            \addlegendentry{TREC}

            \addplot[
                dash pattern=on 6pt off 2pt on 2pt off 2pt,
                mark=triangle*,
                color=color3,
                very thick,
                mark options={scale=1.0, solid},
            ]
            coordinates {
                (0,13.5) (25,17) (50,18) (100,22.5) (200,24)
            };
            \addlegendentry{RTE}

            \addplot[
                densely dashdotted,
                mark=diamond*,
                color=color4,
                very thick,
                mark options={scale=1.0, solid},
            ]
            coordinates {
                (0,12.5) (25,50.5) (50,51.5) (100,53.5) (200,53)
            };
            \addlegendentry{DBpedia}
            \end{axis}
        \end{tikzpicture}
    \end{subfigure}
    \begin{subfigure}[b]{0.48\linewidth}
        \centering
        \vspace{3em}
        \begin{tikzpicture}
            \begin{axis}[
                width=\linewidth,
                height=\linewidth,
                xlabel={\scriptsize Number of Shots ($k$-shot)},
                ylabel={\scriptsize Exact Matches (\%)},
                ylabel style={yshift=-1.4em},
                xlabel style={yshift=0.4em},
                xmin=-15, xmax=215,
                ymin=-5, ymax=75,
                ytick={0,10,20,30,40,50,60,70},
                legend pos=north west,
                legend cell align={left},
                grid=major,
                xtick={0,25,50,100,200},
                xticklabels={0,25,50,100,200},
                extra x tick labels={$\vert\vert$},
                extra x tick style={grid=none},
                tick label style={font=\scriptsize},
                x tick label style={font=\scriptsize},
                axis line style={line width=0.65pt},
                legend style={
                    font=\scriptsize,
                    nodes={scale=0.45, transform shape},
                    fill=white, fill opacity=0.3, text opacity=1,
                    draw opacity=0.6, draw=white!50!black,
                    rounded corners=2pt
                },
                legend image post style={opacity=1},
            ]
            \addplot[
                densely dashed,
                mark=*,
                color=color1,
                very thick,
                mark options={scale=0.8, solid},
            ]
            coordinates {
                (0,0.5) (25,40.5) (50,47) (100,53) (200,63)
            };
            \addlegendentry{WNLI}

            \addplot[
                densely dotted,
                mark=square*,
                color=color2,
                very thick,
                mark options={scale=0.73, solid},
            ]
            coordinates {
                (0,11.5) (25,32.5) (50,30) (100,30.5) (200,31)
            };
            \addlegendentry{TREC}

            \addplot[
                dash pattern=on 6pt off 2pt on 2pt off 2pt,
                mark=triangle*,
                color=color3,
                very thick,
                mark options={scale=1.0, solid},
            ]
            coordinates {
                (0,1) (25,5.5) (50,5.5) (100,6) (200,8)
            };
            \addlegendentry{RTE}

            \addplot[
                densely dashdotted,
                mark=diamond*,
                color=color4,
                very thick,
                mark options={scale=1.0, solid},
            ]
            coordinates {
                (0,0) (25,22) (50,24.5) (100,28.5) (200,27.5)
            };
            \addlegendentry{DBpedia}
            \end{axis}
        \end{tikzpicture}
    \end{subfigure}
    
    \vspace{-0.2cm}
    
    \begin{center}
        \parbox{0.5\textwidth}{\centering \small\textbf{(3) Only Segment Pairs}}
    \end{center}
    \vspace{-1.1cm}
    
    \begin{subfigure}[b]{0.48\linewidth}
        \centering
        \begin{tikzpicture}
            \begin{axis}[
                width=\linewidth,
                height=\linewidth,
                xlabel={\scriptsize Number of Shots ($k$-shot)},
                ylabel={\tiny Exact \& Near-Exact Matches (\%)},
                ylabel style={yshift=-1.4em},
                xlabel style={yshift=0.4em},
                xmin=-15, xmax=215,
                ymin=-5, ymax=85,
                ytick={0,10,20,30,40,50,60,70,80},
                legend pos=north east,
                legend cell align={left},
                grid=major,
                xtick={0,25,50,100,200},
                xticklabels={0,25,50,100,200},
                extra x tick labels={$\vert\vert$},
                extra x tick style={grid=none},
                tick label style={font=\scriptsize},
                x tick label style={font=\scriptsize},
                axis line style={line width=0.65pt},
                legend style={
                    font=\scriptsize,
                    nodes={scale=0.45, transform shape},
                    fill=white, fill opacity=0.3, text opacity=1,
                    draw opacity=0.6, draw=white!50!black,
                    rounded corners=2pt
                },
                legend image post style={opacity=1},
            ]
            \addplot[
                densely dashed,
                mark=*,
                color=color1,
                very thick,
                mark options={scale=0.8, solid},
            ]
            coordinates {
                (0,8) (25,40.5) (50,41.5) (100,47) (200,52.5)
            };
            \addlegendentry{WNLI}

            \addplot[
                densely dotted,
                mark=square*,
                color=color2,
                very thick,
                mark options={scale=0.73, solid},
            ]
            coordinates {
                (0,14) (25,38) (50,40) (100,39.5) (200,40)
            };
            \addlegendentry{TREC}

            \addplot[
                dash pattern=on 6pt off 2pt on 2pt off 2pt,
                mark=triangle*,
                color=color3,
                very thick,
                mark options={scale=1.0, solid},
            ]
            coordinates {
                (0,10) (25,13.5) (50,16.5) (100,19.5) (200,24)
            };
            \addlegendentry{RTE}

            \addplot[
                densely dashdotted,
                mark=diamond*,
                color=color4,
                very thick,
                mark options={scale=1.0, solid},
            ]
            coordinates {
                (0,23.5) (25,50) (50,49) (100,51) (200,52)
            };
            \addlegendentry{DBpedia}
            \end{axis}
        \end{tikzpicture}
    \end{subfigure}
    \begin{subfigure}[b]{0.48\linewidth}
        \centering
        \vspace{3em}
        \begin{tikzpicture}
            \begin{axis}[
                width=\linewidth,
                height=\linewidth,
                xlabel={\scriptsize Number of Shots ($k$-shot)},
                ylabel={\scriptsize Exact Matches (\%)},
                ylabel style={yshift=-1.4em},
                xlabel style={yshift=0.4em},
                xmin=-15, xmax=215,
                ymin=-5, ymax=75,
                ytick={0,10,20,30,40,50,60,70},
                legend pos=north west,
                legend cell align={left},
                grid=major,
                xtick={0,25,50,100,200},
                xticklabels={0,25,50,100,200},
                extra x tick labels={$\vert\vert$},
                extra x tick style={grid=none},
                tick label style={font=\scriptsize},
                x tick label style={font=\scriptsize},
                axis line style={line width=0.65pt},
                legend style={
                    font=\scriptsize,
                    nodes={scale=0.45, transform shape},
                    fill=white, fill opacity=0.3, text opacity=1,
                    draw opacity=0.6, draw=white!50!black,
                    rounded corners=2pt
                },
                legend image post style={opacity=1},
            ]
            \addplot[
                densely dashed,
                mark=*,
                color=color1,
                very thick,
                mark options={scale=0.8, solid},
            ]
            coordinates {
                (0,0) (25,29) (50,32) (100,39.5) (200,48)
            };
            \addlegendentry{WNLI}

            \addplot[
                densely dotted,
                mark=square*,
                color=color2,
                very thick,
                mark options={scale=0.73, solid},
            ]
            coordinates {
                (0,1) (25,28.5) (50,29) (100,29) (200,29.5)
            };
            \addlegendentry{TREC}

            \addplot[
                dash pattern=on 6pt off 2pt on 2pt off 2pt,
                mark=triangle*,
                color=color3,
                very thick,
                mark options={scale=1.0, solid},
            ]
            coordinates {
                (0,0) (25,3.5) (50,5) (100,5.5) (200,8)
            };
            \addlegendentry{RTE}

            \addplot[
                densely dashdotted,
                mark=diamond*,
                color=color4,
                very thick,
                mark options={scale=1.0, solid},
            ]
            coordinates {
                (0,0) (25,20.5) (50,22.5) (100,24) (200,22)
            };
            \addlegendentry{DBpedia}
            \end{axis}
        \end{tikzpicture}
    \end{subfigure}

    \caption{\textbf{Results on quantifying memorization in different ICL regimes for all settings.}
    Plots on the left display memorization using \textit{exact and near‐exact matches}; 
    plots on the right show \textit{only exact matches}. GPT-4 is the underlying model.}
    \label{fig:memorization-quantification}
\end{wrapfigure}
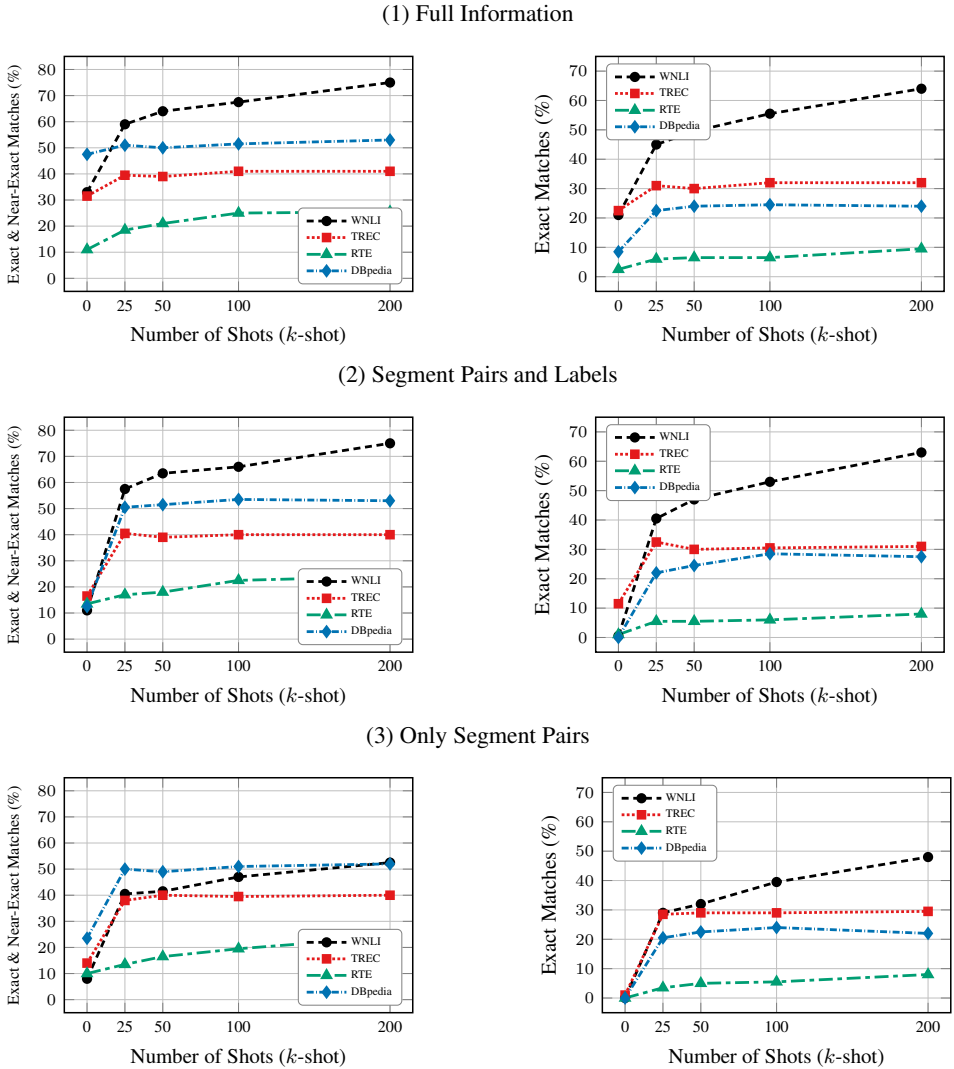

Figure \ref{fig:memorization-quantification} presents a series of plots that quantify memorization across various ICL regimes in our three settings: (1) full information, (2) segment pairs and labels, and (3) only segment pairs.
In this figure, memorization is represented using both exact and near-exact matches (left-hand plots), as well as exact matches alone (right-hand plots). \textit{Unless otherwise stated, all our observations are based on memorization quantified by both exact and near-exact matches.}
We use the first setting only for comparing with maximum memorization, the second setting when discussing memorization in ICL (as it closely resembles standard ICL), and the third setting to identify the key element contributing the most to memorization in ICL. 

We draw six observations on memorization in ICL:

\textbf{Observation 1:} ICL significantly surfaces memorization compared to zero-shot learning. For instance, in the left plot under the segment pairs and labels setting in Figure \ref{fig:memorization-quantification}, memorization in zero-shot regimes ranges from 11\%--16.50\%. This increases to 18\%--63.50\% in few-shot regimes and further to 24\%--75\% in many-shot, more than doubling the amount in zero-shot. 

\textbf{Observation 2:} In terms of memorization behavior, providing only a few demonstrations (e.g., 25 shots) \textit{sharply} increases memorization in ICL for most datasets. While this increase continues for larger shots in some datasets, it plateaus for others. For example, in the aforementioned plot, for the WNLI and RTE datasets, memorization increases up to 75\% and 24\% at 200-shot. However, for the TREC and DBpedia datasets, it levels of at around 40\% and 53\%, respectively, after 25-shot.

\textbf{Observation 3:} Memorization tends to remain stable across many-shot regimes for most datasets. However, within these regimes, memorization becomes more explicit as near-exact matches gradually transform into exact matches, as shown in Figure \ref{fig:exact-vs-near-exact-matches} in Appendix \ref{app:exact-and-near-exact-matches}. This highlights the importance of near-exact matches in quantifying memorization, as they indeed indicate memorization. Table \ref{tab:turning-near-exact-atches-to-exact-matches} in Appendix \ref{app:exact-and-near-exact-matches} also provides examples of near-exact matches turning into exact matches as the number of demonstrations increases.
In addition, as shown in all plots of Figure \ref{fig:memorization-quantification}, the memorization pattern remains consistent whether quantified by exact and near-exact matches together or by exact matches alone, further validating that near-exact matches signal memorization.

\textbf{Observation 4:} As shown in the left-hand plots of Figure \ref{fig:memorization-quantification}, the amount of surfaced memorization in the setting with segment pairs and labels reaches its maximum---equivalent to the full information setting---as soon as a few demonstrations (25 shots and more) are provided in the input prompt. In other words, the key difference between the full information setting and the setting with segment pairs and labels lies in the zero-shot regimes. In fact, the \textit{dataset-specific information is overshadowed by the information from segment pairs and labels (or demonstrations) in the input prompt in terms of contributing to the memorization in ICL.} 

\textbf{Observation 5:} Individual instances of the same context can significantly increase memorization. This is supported by viewing demonstrations as individual dataset instances within the same dataset (context) that contribute to memorization, as seen in all plots of Figure \ref{fig:memorization-quantification}. This complements the findings of \citet{DBLP:conf/iclr/CarliniIJLTZ23} that memorization significantly increases with the tokens of context used to prompt the model.\footnote{See Appendix \ref{sec:comparing-observations-with-previous-studies} for further discussion.}


\textbf{Observation 6:} Comparing memorization levels across three settings (all left-hand plots of Figure \ref{fig:memorization-quantification}) shows that maximum memorization in ICL can be achieved even when \textit{only segment pairs} are included in the input prompt. \textit{This indicates that demonstrations without their respective labels, i.e., the segment pairs alone, are key element contributing to memorization in ICL.}

To clarify, the WNLI dataset is not an exception to this observation, although it experiences a slightly larger decrease (e.g., 22.50\% in the 200-shot regime) compared to other datasets. We discovered that in WNLI dataset, multiple sentence pairs share sentence 1 with different labels. For instance, for \textit{``Bill passed the gameboy to John because his turn was next.''} as sentence 1, there are different options for sentence 2, such as \textit{``John's turn was next.''} and \textit{``Bill's turn was next.''}, with distinct labels—\textit{entailment} and \textit{not entailment}, respectively. Hence, when the model is prompted with only sentence 1 to generate sentence 2, the completion could be either of these options. However, our criteria for exact and near-exact matches require both semantic and structural similarity. If the completion does not semantically or structurally match the original sentence 2, even if it matches the other sentence 2 with a different label, it is not counted as an exact/near-exact match. This approach is consistent with prior work on memorization in LLMs \citep{DBLP:conf/iclr/CarliniIJLTZ23}. We found several cases of this occurring in WNLI dataset, and the number of such cases exactly corresponds with the observed drop in the setting containing only segment pairs.



\subsection{Results on Performance and Memorization}
\label{subsec:performance-and-memorization}

\begin{table}[!]
\centering
\scriptsize
\begin{minipage}[!ht]{0.47\linewidth}
    \centering
    \adjustbox{valign=t}{%
    \begin{minipage}{\linewidth}
    \fontsize{8.5pt}{10pt}\selectfont
    \renewcommand{\arraystretch}{0.9}
    \captionof{table}{Pearson correlation between overall performance and memorization across all settings. Here, memorization is quantified using \textit{both exact and near-exact matches.}}
    \label{tab:pearson-values-for-exact-and-near-exact-matches}
    \begin{tabularx}{\linewidth}{@{\hskip 0pt}l@{\hskip 3.5pt}c@{\hskip 3.5pt}c@{\hskip 3.5pt}c@{\hskip 3.5pt}c@{\hskip 3.5pt}}

        \toprule
        \textbf{Setting} & \textbf{WNLI} & \textbf{TREC} & \textbf{DBpedia} & \textbf{RTE} \\
        \midrule
        Full Information       & 0.98 & 0.95 & 0.91 & --0.55 \\
        Seg. Pairs \& Labels & 1.00 & 0.91 & 0.88 & --0.30 \\
        Only Seg. Pairs      & 0.99 & 0.92 & 0.89 & --0.30 \\
        \bottomrule
    \end{tabularx}
    \end{minipage}%
    }
\end{minipage}
\hspace{0.04\linewidth}
\begin{minipage}[!ht]{0.47\linewidth}
    \centering
    \adjustbox{valign=t}{%
    \begin{minipage}{\linewidth}
    \fontsize{8.5pt}{10pt}\selectfont
    \renewcommand{\arraystretch}{0.9}
    \captionof{table}{Pearson correlation between overall performance and memorization across all settings. Here, memorization is quantified using \textit{only exact matches}.}
    \label{tab:pearson-values-for-only-exact-matches}
    \begin{tabularx}{\linewidth}{@{\hskip 0pt}l@{\hskip 3.5pt}c@{\hskip 3.5pt}c@{\hskip 3.5pt}c@{\hskip 3.5pt}c@{\hskip 3.5pt}}

        \toprule
        \textbf{Setting} & \textbf{WNLI} & \textbf{TREC} & \textbf{DBpedia} & \textbf{RTE} \\
        \midrule
        Full Information       & 0.97 & 0.87 & 0.88 & --0.40 \\
        Seg. Pairs \& Labels & 0.99 & 0.77 & 0.93 & --0.50 \\
        Only Seg. Pairs      & 0.97 & 0.82 & 0.89 & --0.47 \\
        \bottomrule
    \end{tabularx}
    \end{minipage}%
    }
\end{minipage}
\end{table}

\begin{wrapfigure}[48]{r}{0.66\textwidth}
    \centering
    \vspace{-\baselineskip}
    \centering
    
    \begin{center}
        \parbox{0.5\textwidth}{\centering \small\textbf{WNLI}}
    \end{center}
    \vspace{-0.19cm}

    \begin{subfigure}[b]{0.48\linewidth}
        \centering
        \begin{tikzpicture}
            \begin{axis}[
                xlabel={\scriptsize{Number of Shots ($k$-shot)}},
                ylabel={\scriptsize{Accuracy (\%)}},
                ylabel style={yshift=-1.4em},
                xlabel style={yshift=0.4em},
                xmin=-15, xmax=215,
                ymin=73, ymax=103,
                ytick={70, 80, 90, 100},
                legend pos= south east,
                legend cell align={left},
                grid=major,
                xtick={0, 25, 50, 100, 200},
                xticklabels={0, 25, 50, 100, 200},
                extra x tick labels={$\vert\vert$},
                extra x tick style={grid=none},
                tick label style={font=\scriptsize},
                x tick label style={font=\scriptsize},
                width=\linewidth,
                height=\linewidth,
                axis line style={line width=0.65pt}, 
                legend style={font=\scriptsize, nodes={scale=0.45, transform shape}, fill=white, fill opacity=0.3, text opacity=1, draw opacity=0.6, draw=white!50!black, rounded corners=2pt},
                legend image post style={opacity=1},
            ]

            \addplot[
            dash pattern = on 6pt off 2pt on 2pt off 2pt,
            mark=*,
            color=color5,
            very thick, 
            mark options={scale=0.8, solid},
            ]
            coordinates {
            

            (0, 83.75) (25, 90) (50, 90.75) (100, 90.5) (200, 92)

            };
            \addlegendentry{Overall Performance}
            
            \addplot[
            densely dashdotted,
            mark=diamond*,
            color=color6,
            very thick, 
            mark options={scale=1.0, solid},
            ]
            coordinates {
            
            (0, 90.91) (25, 95.65) (50, 94.1) (100, 94.7) (200, 95.33)
            };
            \addlegendentry{Perf. for Memorized Instances}

            \addplot[
            densely dotted,
            mark=triangle*,
            color=color7,
            very thick, 
            mark options={scale=1.0, solid},
            ]
            coordinates {

            (0, 82.86) (25, 83.35) (50, 84.93) (100, 82.35) (200, 82)

            };
            \addlegendentry{Perf. for Non-Memorized Instances}

            \end{axis}
        \end{tikzpicture}
        \label{subfig:performance-on-wnli}
    \end{subfigure}
    \hfill
    \begin{subfigure}[b]{0.48\linewidth}
        \centering
        \begin{tikzpicture}
            \begin{axis}[
                xlabel={\scriptsize{Number of Shots ($k$-shot)}},
                ylabel={\tiny{Exact \& Near-Exact Matches (\%)}},
                ylabel style={yshift=-1.4em},
                xlabel style={yshift=0.4em},
                xmin=-15, xmax=215,
                ymin=-5, ymax=85,
                ytick={0, 10, 20, 30, 40, 50, 60, 70, 80},
                legend pos=south east,
                legend cell align={left},
                grid=major,
                xtick={0, 25, 50, 100, 200},
                xticklabels={0, 25, 50, 100, 200},
                extra x tick labels={$\vert\vert$},
                extra x tick style={grid=none},
                tick label style={font=\scriptsize},
                x tick label style={font=\scriptsize},
                width=\linewidth,
                height=\linewidth,
                axis line style={line width=0.65pt}, 
                legend style={font=\scriptsize, nodes={scale=0.45, transform shape}, fill=white, fill opacity=0.3, text opacity=1, draw opacity=0.6, draw=white!50!black, rounded corners=2pt},
                legend image post style={opacity=1}
            ]
            \addplot[
            dash pattern = on 6pt off 2pt on 2pt off 2pt,
            mark=*,
            color=color1,
            very thick,
            mark options={scale=0.8, solid},
            ]
            coordinates {
            (0, 33) (25, 59) (50, 64) (100, 67.5) (200, 75)
            };
            \addlegendentry{Full Information}

            \addplot[
            densely dotted,
            mark=diamond*,
            color=color2,
            very thick,
            mark options={scale=1.0, solid},
            ]
            coordinates {
            (0, 11) (25, 57.5) (50, 63.5) (100, 66) (200, 75)
            };
            \addlegendentry{Segment Pairs \& Labels}

            \addplot[
            densely dashdotted,
            mark=triangle*,
            color=color4,
            very thick, 
            mark options={scale=1.0, solid},
            ]
            coordinates {
            (0, 8) (25, 40.5) (50, 41.5) (100, 47) (200, 52.5)
            };
            \addlegendentry{Only Segment Pairs}
            \end{axis}
        \end{tikzpicture}
        \label{subfig:memorization-on-wnli}
    \end{subfigure}
    \label{fig:wnli-performance-vs-memorization}
\vspace{-0.3cm}
    \centering
    
    \begin{center}
        \parbox{0.5\textwidth}{\centering \small\textbf{TREC}}
    \end{center}
    \vspace{-0.19cm}

    \begin{subfigure}[b]{0.48\linewidth}
        \centering
        \begin{tikzpicture}
            \begin{axis}[
                xlabel={\scriptsize{Number of Shots ($k$-shot)}},
                ylabel={\scriptsize{Accuracy (\%)}},
                ylabel style={yshift=-1.4em},
                xlabel style={yshift=0.4em},
                xmin=-15, xmax=215,
                ymin=67, ymax=103,
                ytick={70, 80, 90, 100},
                legend pos= south east,
                legend cell align={left},
                grid=major,
                xtick={0, 25, 50, 100, 200},
                xticklabels={0, 25, 50, 100, 200},
                extra x tick labels={$\vert\vert$},
                extra x tick style={grid=none},
                tick label style={font=\scriptsize},
                x tick label style={font=\scriptsize},
                width=\linewidth,
                height=\linewidth,
                axis line style={line width=0.65pt}, 
                legend style={font=\scriptsize, nodes={scale=0.45, transform shape}, fill=white, fill opacity=0.3, text opacity=1, draw opacity=0.6, draw=white!50!black, rounded corners=2pt},
                legend image post style={opacity=1},
            ]

            \addplot[
             dash pattern = on 6pt off 2pt on 2pt off 2pt,
            mark=*,
            color=color5,
            very thick, 
            mark options={scale=0.8, solid},
            ]
            coordinates {
            (0, 83.75) (25, 86.5) (50, 88.5) (100, 90.5) (200, 90.75)
            };
            \addlegendentry{Overall Performance}
            
            \addplot[
            densely dashdotted,
            mark=diamond*,
            color=color6,
            very thick, 
            mark options={scale=1.0, solid},
            ]
            coordinates {
            (0, 100.00) (25, 95.68) (50, 93.59) (100, 94.37) (200, 93.75)
            };
            \addlegendentry{Perf. for Memorized Instances}

            \addplot[
            densely dotted,
            mark=triangle*,
            color=color7,
            very thick, 
            mark options={scale=1.0, solid},
            ]
            coordinates {
            (0, 80.54) (25, 80.25) (50, 85.25) (100, 87.91) (200, 88.75)
            };
            \addlegendentry{Perf. for Non-Memorized Instances}

            \end{axis}
        \end{tikzpicture}
        \label{subfig:performance-on-trec}
    \end{subfigure}
    \hfill
    \begin{subfigure}[b]{0.48\linewidth}
        \centering
        \begin{tikzpicture}
            \begin{axis}[
                xlabel={\scriptsize{Number of Shots ($k$-shot)}},
                ylabel={\tiny{Exact \& Near-Exact Matches (\%)}},
                ylabel style={yshift=-1.4em},
                xlabel style={yshift=0.4em},
                xmin=-15, xmax=215,
                ymin=5, ymax=55,
                ytick={0, 10, 20, 30, 40, 50, 60, 70},
                legend pos=south east,
                legend cell align={left},
                grid=major,
                xtick={0, 25, 50, 100, 200},
                xticklabels={0, 25, 50, 100, 200},
                extra x tick labels={$\vert\vert$},
                extra x tick style={grid=none},
                tick label style={font=\scriptsize},
                x tick label style={font=\scriptsize},
                width=\linewidth,
                height=\linewidth,
                axis line style={line width=0.65pt}, 
                legend style={font=\scriptsize, nodes={scale=0.45, transform shape}, fill=white, fill opacity=0.3, text opacity=1, draw opacity=0.6, draw=white!50!black, rounded corners=2pt},
                legend image post style={opacity=1}
            ]
        \addplot[
            dash pattern = on 6pt off 2pt on 2pt off 2pt,
            mark=*,
            color=color1,
            very thick,
            mark options={scale=0.8, solid},
            ]
            coordinates {
             (0, 31.5) (25, 39.5) (50, 39) (100, 41) (200, 41)
            };
            \addlegendentry{Full Information}

        \addplot[
            densely dotted,
            mark=diamond*,
            color=color2,
            very thick, 
            mark options={scale=1.0, solid},
            ]
            coordinates {
            (0, 16.5) (25, 40.5) (50, 39) (100, 40) (200, 40)
            };
            \addlegendentry{Segment Pairs \& Labels}

        \addplot[
            densely dashdotted,
            mark=triangle*,
            color=color4,
            very thick, 
            mark options={scale=1.0, solid},
            ]
            coordinates {
            (0, 14) (25, 38) (50, 40) (100, 39.5) (200, 40)
            };
            \addlegendentry{Only Segment Pairs}
            \end{axis}
        \end{tikzpicture}
        \label{subfig:memorization-on-trec}
    \end{subfigure}
    \label{fig:trec-performance-vs-memorization}
\vspace{-0.3cm}
    \centering
    
    \begin{center}
        \parbox{0.5\textwidth}{\centering \small\textbf{DBpedia}}
    \end{center}
    \vspace{-0.19cm}

    \begin{subfigure}[b]{0.48\linewidth}
        \centering
        \begin{tikzpicture}
            \begin{axis}[
                xlabel={\scriptsize{Number of Shots ($k$-shot)}},
                ylabel={\scriptsize{Accuracy (\%)}},
                ylabel style={yshift=-1.4em},
                xlabel style={yshift=0.4em},
                xmin=-15, xmax=215,
                ymin=88, ymax=102,
                ytick={70, 80, 90, 95, 100},
                legend pos= south east,
                legend cell align={left},
                grid=major,
                xtick={0, 25, 50, 100, 200},
                xticklabels={0, 25, 50, 100, 200},
                extra x tick labels={$\vert\vert$},
                extra x tick style={grid=none},
                tick label style={font=\scriptsize},
                x tick label style={font=\scriptsize},
                width=\linewidth,
                height=\linewidth,
                axis line style={line width=0.65pt}, 
                legend style={font=\scriptsize, nodes={scale=0.45, transform shape}, fill=white, fill opacity=0.3, text opacity=1, draw opacity=0.6, draw=white!50!black, rounded corners=2pt},
                legend image post style={opacity=1},
            ]

            \addplot[
             dash pattern = on 6pt off 2pt on 2pt off 2pt,
            mark=*,
            color=color5,
            very thick, 
            mark options={scale=0.8, solid},
            ]
            coordinates {
            (0, 96) (25, 97.75) (50, 97.5) (100, 99) (200, 98.5)
            };
            \addlegendentry{Overall Performance}

            \addplot[
            densely dashdotted,
            mark=diamond*,
            color=color6,
            very thick, 
            mark options={scale=1.0, solid},
            ]
            coordinates {
            (0, 100.00) (25, 99.01) (50, 99.03) (100, 100.00) (200, 98.11)
            };
            \addlegendentry{Perf. for Memorized Instances}

            \addplot[
            densely dotted,
            mark=triangle*,
            color=color7,
            very thick, 
            mark options={scale=1.0, solid},
            ]
            coordinates {
            (0, 95.43) (25, 96.46) (50, 95.88) (100, 97.85) (200, 98.94)
            };
            \addlegendentry{Perf. for Non-Memorized Instances}

            \end{axis}
        \end{tikzpicture}
        \label{subfig:performance-on-dbpedia}
    \end{subfigure}
    \hfill
    \begin{subfigure}[b]{0.48\linewidth}
        \centering
        \begin{tikzpicture}
            \begin{axis}[
                xlabel={\scriptsize{Number of Shots ($k$-shot)}},
                ylabel={\tiny{Exact \& Near-Exact Matches (\%)}},
                ylabel style={yshift=-1.4em},
                xlabel style={yshift=0.4em},
                xmin=-15, xmax=215,
                ymin=4, ymax=64,
                ytick={0, 10, 20, 30, 40, 50, 60, 70},
                legend pos=south east,
                legend cell align={left},
                grid=major,
                xtick={0, 25, 50, 100, 200},
                xticklabels={0, 25, 50, 100, 200},
                extra x tick labels={$\vert\vert$},
                extra x tick style={grid=none},
                tick label style={font=\scriptsize},
                x tick label style={font=\scriptsize},
                width=\linewidth,
                height=\linewidth,
                axis line style={line width=0.65pt}, 
                legend style={font=\scriptsize, nodes={scale=0.45, transform shape}, fill=white, fill opacity=0.3, text opacity=1, draw opacity=0.6, draw=white!50!black, rounded corners=2pt},
                legend image post style={opacity=1}
            ]
        \addplot[
            dash pattern = on 6pt off 2pt on 2pt off 2pt,
            mark=*,
            color=color1,
            very thick,
            mark options={scale=0.8, solid},
            ]
            coordinates {
            (0, 47.5) (25, 51) (50, 50) (100, 51.5) (200, 53)
            };
            \addlegendentry{Full Information}

        \addplot[
            densely dotted,
            mark=diamond*,
            color=color2,
            very thick, 
            mark options={scale=1.0, solid},
            ]
            coordinates {
            (0, 12.5) (25, 50.5) (50, 51.5) (100, 53.5) (200, 53)
            };
            \addlegendentry{Segment Pairs \& Labels}

        \addplot[
            densely dashdotted,
            mark=triangle*,
            color=color4,
            very thick, 
            mark options={scale=1.0, solid},
            ]
            coordinates {
            (0, 23.5) (25, 50) (50, 49) (100, 51) (200, 52)
            };
            \addlegendentry{Only Segment Pairs}
            \end{axis}
        \end{tikzpicture}
        \label{subfig:memorization-on-dbpedia}
    \end{subfigure}
    \label{fig:dbpedia-performance-vs-memorization}
\vspace{-0.3cm}
    \centering
    
    \begin{center}
        \parbox{0.5\textwidth}{\centering \small\textbf{RTE}}
    \end{center}
    \vspace{-0.19cm}
    
    \begin{subfigure}[b]{0.48\linewidth}
        \centering
        \begin{tikzpicture}
            \begin{axis}[
                xlabel={\scriptsize{Number of Shots ($k$-shot)}},
                ylabel={\scriptsize{Accuracy (\%)}},
                ylabel style={yshift=-1.4em},
                xlabel style={yshift=0.4em},
                xmin=-15, xmax=215,
                ymin=76, ymax=103,
                ytick={70, 80, 85, 90, 95, 100},
                legend pos= south east,
                legend cell align={left},
                grid=major,
                xtick={0, 25, 50, 100, 200},
                xticklabels={0, 25, 50, 100, 200},
                extra x tick labels={$\vert\vert$},
                extra x tick style={grid=none},
                tick label style={font=\scriptsize},
                x tick label style={font=\scriptsize},
                width=\linewidth,
                height=\linewidth,
                axis line style={line width=0.65pt}, 
                legend style={font=\scriptsize, nodes={scale=0.45, transform shape}, fill=white, fill opacity=0.3, text opacity=1, draw opacity=0.6, draw=white!50!black, rounded corners=2pt},
                legend image post style={opacity=1},
            ]

            \addplot[
            dash pattern = on 6pt off 2pt on 2pt off 2pt,
            mark=*,
            color=color5,
            very thick, 
            mark options={scale=0.8, solid},
            ]
            coordinates {
            (0, 92.5) (25, 90.5) (50, 88.75) (100, 90) (200, 91.25)
            };
            \addlegendentry{Overall Performance}
            
            \addplot[
            densely dashdotted,
            mark=diamond*,
            color=color6,
            very thick, 
            mark options={scale=1.0, solid},
            ]
            coordinates {
            (0, 92.59) (25, 98.53) (50, 97.22) (100, 93.33) (200, 92.71)
            };
            \addlegendentry{Perf. for Memorized Instances}

            \addplot[
            densely dotted,
            mark=triangle*,
            color=color7,
            very thick, 
            mark options={scale=1.0, solid},
            ]
            coordinates {
            (0, 92.49) (25, 88.85) (50, 86.89) (100, 90.32) (200, 90.79)
            };
            \addlegendentry{Perf. for Non-Memorized Instances}

            \end{axis}
        \end{tikzpicture}
        \label{subfig:performance-on-rte}
    \end{subfigure}
    \hfill
    \begin{subfigure}[b]{0.48\linewidth}
        \centering
        \begin{tikzpicture}
            \begin{axis}[
                xlabel={\scriptsize{Number of Shots ($k$-shot)}},
                ylabel={\tiny{Exact \& Near-Exact Matches (\%)}},
                ylabel style={yshift=-1.4em},
                xlabel style={yshift=0.4em},
                xmin=-15, xmax=215,
                ymin=-4, ymax=34,
                ytick={0, 10, 20, 30, 40, 50, 60, 70},
                legend pos=south east,
                legend cell align={left},
                grid=major,
                xtick={0, 25, 50, 100, 200},
                xticklabels={0, 25, 50, 100, 200},
                extra x tick labels={$\vert\vert$},
                extra x tick style={grid=none},
                tick label style={font=\scriptsize},
                x tick label style={font=\scriptsize},
                width=\linewidth,
                height=\linewidth,
                axis line style={line width=0.65pt}, 
                legend style={font=\scriptsize, nodes={scale=0.45, transform shape}, fill=white, fill opacity=0.3, text opacity=1, draw opacity=0.6, draw=white!50!black, rounded corners=2pt},
                legend image post style={opacity=1}
            ]
        \addplot[
            dash pattern = on 6pt off 2pt on 2pt off 2pt,
            mark=*,
            color=color1,
            very thick,
            mark options={scale=0.8, solid},
            ]
            coordinates {
            (0, 11) (25, 18.5) (50, 21) (100, 25) (200, 25.5)
            };
            \addlegendentry{Full Information}

        \addplot[
            densely dotted,
            mark=diamond*,
            color=color2,
            very thick, 
            mark options={scale=1.0, solid},
            ]
            coordinates {
            (0, 13.5) (25, 17) (50, 18) (100, 22.5) (200, 24)
            };
            \addlegendentry{Segment Pairs \& Labels}

        \addplot[
            densely dashdotted,
            mark=triangle*,
            color=color4,
            very thick, 
            mark options={scale=1.0, solid},
            ]
            coordinates {
            (0, 10) (25, 13.5) (50, 16.5) (100, 19.5) (200, 24)
            };
            \addlegendentry{Only Segment Pairs}
            \end{axis}
        \end{tikzpicture}
    \end{subfigure}
    \caption{\textbf{Performance vs. memorization in different ICL regimes for all settings.} Left plots show performance, and right plots display memorization across all settings. Memorization plots are duplicated from Figure \ref{fig:memorization-quantification} for comparison.}%
    \label{fig:performance-vs-memorization}%
\end{wrapfigure}
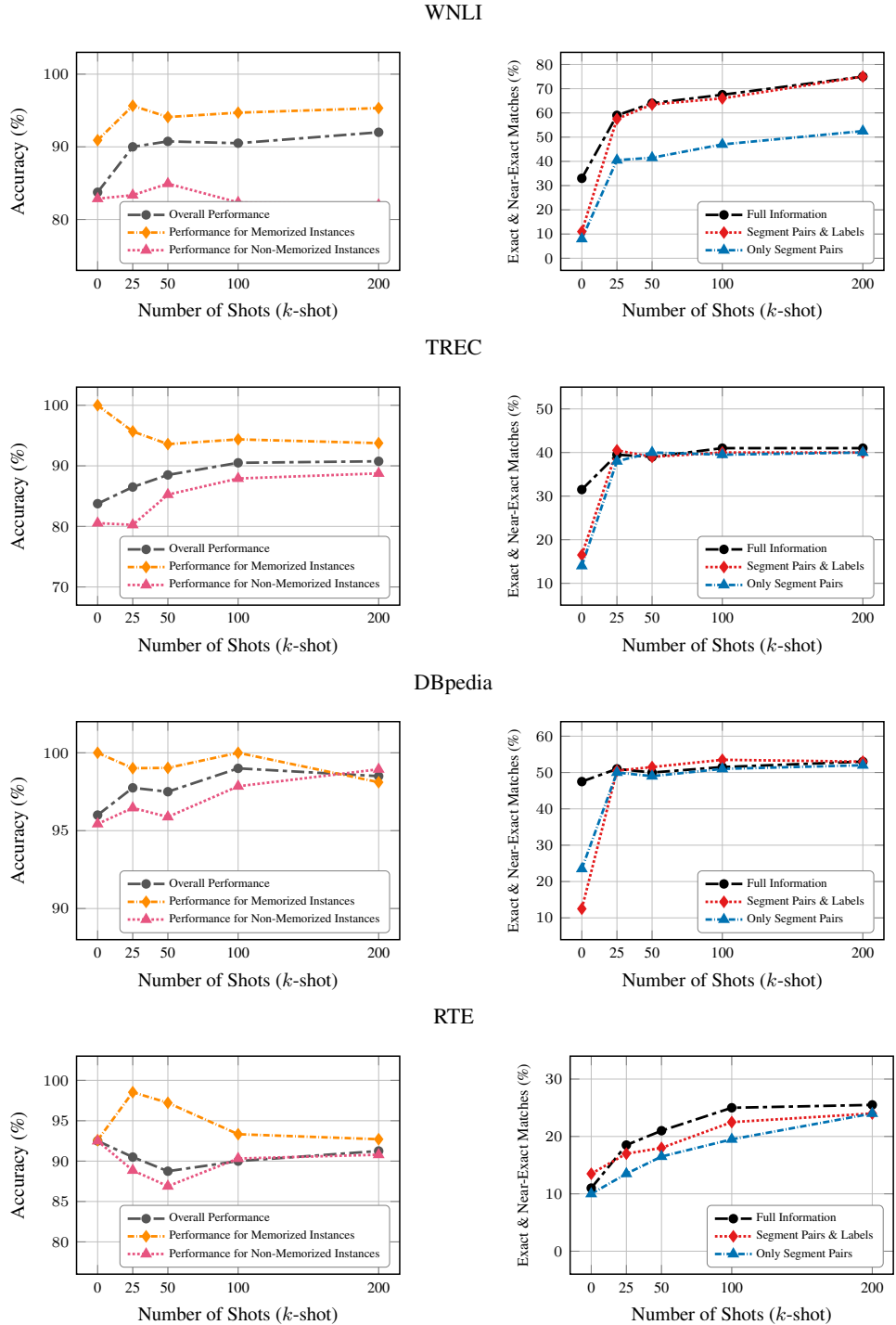

Figure \ref{fig:performance-vs-memorization} presents plots comparing performance (left plots) and memorization (right plots) across different ICL regimes in all three settings. 
Accordingly, Tables \ref{tab:pearson-values-for-exact-and-near-exact-matches} and \ref{tab:pearson-values-for-only-exact-matches} list the Pearson correlation coefficients between overall performance and memorization, with memorization quantified using both exact and near-exact matches, and only exact matches, respectively.

We highlight three key observations regarding the relationship between performance and memorization in ICL:\footnote{See Appendix \ref{app:extended-performance-and-memorization} for an extended discussion.}

\textbf{Observation 1:} ICL outperforms zero-shot learning when the surfaced memorization level in few-shot regimes is substantial, reaching around 40\% or higher. This is evident in the results from the WNLI, TREC, and DBpedia datasets in Figure \ref{fig:performance-vs-memorization}.




\textbf{Observation 2:} Performance on memorized instances is consistently higher than on non-memorized instances across nearly all settings, from zero-shot to many-shot regimes. This can be observed in the left-hand plots of Figure \ref{fig:performance-vs-memorization}.

\textbf{Observation 3:} As indicated in Tables \ref{tab:pearson-values-for-exact-and-near-exact-matches} and \ref{tab:pearson-values-for-only-exact-matches}, \textit{when providing demonstrations in ICL leads to performance improvement compared to zero-shot learning, this is highly correlated with memorization.} Specifically, Pearson coefficients provide very strong evidence of this correlation.



\section{Related Work}

\textbf{In-Context Learning.} Proposed by \citet{brown2020language}, ICL enhances performance in LLMs without additional training by including a few demonstrations in the input prompt. Despite its simplicity, the internal mechanism of ICL is not yet well understood.
Several studies explored ICL from various perspectives: \citet{DBLP:conf/acl/LuBM0S22} examined the impact of the order of demonstrations, \citet{DBLP:conf/emnlp/BolucuR023} investigated the effect of example selection, \citet{DBLP:conf/icml/ZhaoWFK021} explored label, recency, and common token biases in ICL, \citet{DBLP:journals/corr/abs-2307-05052} studied the influence of input distribution and explanations, and \citet{DBLP:conf/emnlp/MinLHALHZ22} looked into the role of labels and found that randomly replacing labels does not harm performance while others showed that this is not true for all tasks and models \citep{DBLP:conf/emnlp/YooKKCJLLK22,DBLP:journals/corr/abs-2307-12375,DBLP:journals/corr/abs-2402-18819}.
Different perspectives were employed to better understand ICL: \citet{DBLP:conf/emnlp/HendelGG23} viewed ICL as compressing the training set into a single task vector to produce output, while \citet{DBLP:conf/icml/OswaldNRSMZV23} interpreted ICL as gradient descent, a view refuted by \citet{DBLP:journals/corr/abs-2311-07772}.
Some research efforts focused on maximizing ICL performance through different paradigms: several studies extremely increased demonstrations in the input prompt \citep{DBLP:journals/corr/abs-2405-00200,DBLP:journals/corr/abs-2404-11018,DBLP:journals/corr/abs-2305-15005,DBLP:journals/corr/abs-2309-10954,DBLP:journals/corr/abs-2312-11805}, \citet{DBLP:conf/naacl/MinLZH22} fine-tuned models to perform ICL, and \citet{DBLP:conf/icml/ZhaoWFK021} used prompt engineering to enhance ICL.
Recent research also revealed additional capabilities of ICL beyond its performance on benchmarks, such as regression \citep{DBLP:journals/corr/abs-2404-07544}, kNN \citep{DBLP:journals/corr/abs-2404-11018,DBLP:conf/nips/DinhZZLGRSP022}, and jailbreaking \citep{anil2024many}.

\textbf{Memorization.} Memorization is a well-studied topic in the context of language models \citep{DBLP:conf/iclr/CarliniIJLTZ23,DBLP:conf/uss/CarliniTWJHLRBS21,DBLP:conf/kdd/SongS19,DBLP:conf/nips/ZhangILJTC23,DBLP:journals/tacl/McCoySL0C23,DBLP:conf/ccs/SongRS17}. While some studies aimed to verify the presence of memorization \citep{DBLP:conf/aies/0002SAKFLP18,DBLP:journals/corr/abs-2006-07490,DBLP:conf/tsd/ThomasADMK20,DBLP:conf/uss/Carlini0EKS19,DBLP:journals/corr/abs-2311-06233}, others attempt to quantify it \citep{DBLP:conf/iclr/CarliniIJLTZ23,DBLP:conf/uss/CarliniTWJHLRBS21}.
Quantifying memorization is typically handled using membership inference attack \citep{DBLP:conf/sp/ShokriSSS17,DBLP:conf/csfw/YeomGFJ18} to reproduce training data from models \citep{DBLP:conf/iclr/CarliniIJLTZ23,DBLP:conf/uss/CarliniTWJHLRBS21,DBLP:journals/corr/abs-2308-08493}. This quantification is conducted for several reasons, including showcasing potential privacy risks \citep{DBLP:journals/corr/abs-2311-17035,DBLP:conf/nips/BidermanPSSAPR23,DBLP:conf/sp/LukasSSTWB23}, addressing copyright infringement \citep{grynbaum2023times,DBLP:conf/emnlp/KaramolegkouLZS23}, and generating factual information \citep{DBLP:conf/acl/LiRZWLVYK23,DBLP:conf/nips/Tay00NBM000GSCM22,DBLP:journals/corr/abs-2204-06031,petroni2019language,DBLP:conf/eacl/HavivCGSGG23}. On the other hand, several works proposed methodologies to mitigate memorization \citep{DBLP:conf/acl/LeeINZECC22} and its implications \citep{DBLP:conf/icml/KandpalWR22,DBLP:journals/corr/abs-2402-09363,DBLP:journals/corr/abs-2402-10892,DBLP:journals/corr/abs-2310-00646}.

To our knowledge, no prior work studied memorization and its correlation with performance in ICL.

\section{Conclusion}

We studied memorization in in-context learning (ICL) and its correlation with downstream performance in large language models (LLMs). Specifically, we proposed a novel technique to quantify this memorization and identify the most effective element in surfacing it within ICL. 
Our key findings are: (1) ICL significantly surfaces memorization compared to zero-shot learning in most cases, with demonstrations---excluding labels---being the most influential element; and (2) when ICL outperforms zero-shot learning, improved performance shows a strong correlation with memorization.
Overall, our findings position memorization as a new factor impacting ICL.
While our research offers a deeper understanding of ICL, it also presents new challenges, particularly on how much LLMs truly learn from demonstrations in ICL versus how much of their success is due to memorization.

\bibliography{colm2025_conference}

\begin{thebibliography}{80}
\providecommand{\natexlab}[1]{#1}
\providecommand{\url}[1]{\texttt{#1}}
\expandafter\ifx\csname urlstyle\endcsname\relax
  \providecommand{\doi}[1]{doi: #1}\else
  \providecommand{\doi}{doi: \begingroup \urlstyle{rm}\Url}\fi

\bibitem[Agarwal et~al.(2024)Agarwal, Singh, Zhang, Bohnet, Chan, Anand, Abbas, Nova, Co{-}Reyes, Chu, Behbahani, Faust, and Larochelle]{DBLP:journals/corr/abs-2404-11018}
Rishabh Agarwal, Avi Singh, Lei~M. Zhang, Bernd Bohnet, Stephanie C.~Y. Chan, Ankesh Anand, Zaheer Abbas, Azade Nova, John~D. Co{-}Reyes, Eric Chu, Feryal M.~P. Behbahani, Aleksandra Faust, and Hugo Larochelle.
\newblock Many-shot in-context learning.
\newblock \emph{CoRR}, abs/2404.11018, 2024.
\newblock \doi{10.48550/ARXIV.2404.11018}.
\newblock URL \url{https://doi.org/10.48550/arXiv.2404.11018}.

\bibitem[AlKhamissi et~al.(2022)AlKhamissi, Li, Celikyilmaz, Diab, and Ghazvininejad]{DBLP:journals/corr/abs-2204-06031}
Badr AlKhamissi, Millicent Li, Asli Celikyilmaz, Mona~T. Diab, and Marjan Ghazvininejad.
\newblock A review on language models as knowledge bases.
\newblock \emph{CoRR}, abs/2204.06031, 2022.
\newblock \doi{10.48550/ARXIV.2204.06031}.
\newblock URL \url{https://doi.org/10.48550/arXiv.2204.06031}.

\bibitem[Anil et~al.(2024)Anil, Durmus, Sharma, Benton, Kundu, Batson, Rimsky, Tong, Mu, Ford, et~al.]{anil2024many}
Cem Anil, Esin Durmus, Mrinank Sharma, Joe Benton, Sandipan Kundu, Joshua Batson, Nina Rimsky, Meg Tong, Jesse Mu, Daniel Ford, et~al.
\newblock Many-shot jailbreaking.
\newblock \emph{Anthropic, April}, 2024.

\bibitem[Anil et~al.(2023)Anil, Borgeaud, Wu, Alayrac, Yu, Soricut, Schalkwyk, Dai, Hauth, Millican, Silver, Petrov, Johnson, Antonoglou, Schrittwieser, Glaese, Chen, Pitler, Lillicrap, Lazaridou, Firat, Molloy, Isard, Barham, Hennigan, Lee, Viola, Reynolds, Xu, Doherty, Collins, Meyer, Rutherford, Moreira, Ayoub, Goel, Tucker, Piqueras, Krikun, Barr, Savinov, Danihelka, Roelofs, White, Andreassen, von Glehn, Yagati, Kazemi, Gonzalez, Khalman, Sygnowski, and et~al.]{DBLP:journals/corr/abs-2312-11805}
Rohan Anil, Sebastian Borgeaud, Yonghui Wu, Jean{-}Baptiste Alayrac, Jiahui Yu, Radu Soricut, Johan Schalkwyk, Andrew~M. Dai, Anja Hauth, Katie Millican, David Silver, Slav Petrov, Melvin Johnson, Ioannis Antonoglou, Julian Schrittwieser, Amelia Glaese, Jilin Chen, Emily Pitler, Timothy~P. Lillicrap, Angeliki Lazaridou, Orhan Firat, James Molloy, Michael Isard, Paul~Ronald Barham, Tom Hennigan, Benjamin Lee, Fabio Viola, Malcolm Reynolds, Yuanzhong Xu, Ryan Doherty, Eli Collins, Clemens Meyer, Eliza Rutherford, Erica Moreira, Kareem Ayoub, Megha Goel, George Tucker, Enrique Piqueras, Maxim Krikun, Iain Barr, Nikolay Savinov, Ivo Danihelka, Becca Roelofs, Ana{\"{\i}}s White, Anders Andreassen, Tamara von Glehn, Lakshman Yagati, Mehran Kazemi, Lucas Gonzalez, Misha Khalman, Jakub Sygnowski, and et~al.
\newblock Gemini: {A} family of highly capable multimodal models.
\newblock \emph{CoRR}, abs/2312.11805, 2023.
\newblock \doi{10.48550/ARXIV.2312.11805}.
\newblock URL \url{https://doi.org/10.48550/arXiv.2312.11805}.

\bibitem[Anthropic(2024)]{noauthor_introducing_nodate}
Anthropic.
\newblock Introducing {Claude} 3.5 {Sonnet}, 2024.
\newblock URL \url{https://www.anthropic.com/news/claude-3-5-sonnet}.

\bibitem[Askell et~al.(2021)Askell, Bai, Chen, Drain, Ganguli, Henighan, Jones, Joseph, Mann, DasSarma, et~al.]{askell2021general}
Amanda Askell, Yuntao Bai, Anna Chen, Dawn Drain, Deep Ganguli, Tom Henighan, Andy Jones, Nicholas Joseph, Ben Mann, Nova DasSarma, et~al.
\newblock A general language assistant as a laboratory for alignment.
\newblock \emph{arXiv preprint arXiv:2112.00861}, 2021.

\bibitem[Bentivogli et~al.(2009)Bentivogli, Clark, Dagan, and Giampiccolo]{bentivogli2009fifth}
Luisa Bentivogli, Peter Clark, Ido Dagan, and Danilo Giampiccolo.
\newblock The fifth pascal recognizing textual entailment challenge.
\newblock \emph{TAC}, 7:\penalty0 8, 2009.

\bibitem[Bertsch et~al.(2024)Bertsch, Ivgi, Alon, Berant, Gormley, and Neubig]{DBLP:journals/corr/abs-2405-00200}
Amanda Bertsch, Maor Ivgi, Uri Alon, Jonathan Berant, Matthew~R. Gormley, and Graham Neubig.
\newblock In-context learning with long-context models: An in-depth exploration.
\newblock \emph{CoRR}, abs/2405.00200, 2024.
\newblock \doi{10.48550/ARXIV.2405.00200}.
\newblock URL \url{https://doi.org/10.48550/arXiv.2405.00200}.

\bibitem[Biderman et~al.(2023)Biderman, Prashanth, Sutawika, Schoelkopf, Anthony, Purohit, and Raff]{DBLP:conf/nips/BidermanPSSAPR23}
Stella Biderman, USVSN~Sai Prashanth, Lintang Sutawika, Hailey Schoelkopf, Quentin Anthony, Shivanshu Purohit, and Edward Raff.
\newblock Emergent and predictable memorization in large language models.
\newblock In Alice Oh, Tristan Naumann, Amir Globerson, Kate Saenko, Moritz Hardt, and Sergey Levine (eds.), \emph{Advances in Neural Information Processing Systems 36: Annual Conference on Neural Information Processing Systems 2023, NeurIPS 2023, New Orleans, LA, USA, December 10 - 16, 2023}, 2023.
\newblock URL \url{http://papers.nips.cc/paper\_files/paper/2023/hash/59404fb89d6194641c69ae99ecdf8f6d-Abstract-Conference.html}.

\bibitem[B{\"{o}}l{\"{u}}c{\"{u}} et~al.(2023)B{\"{o}}l{\"{u}}c{\"{u}}, Rybinski, and Wan]{DBLP:conf/emnlp/BolucuR023}
Necva B{\"{o}}l{\"{u}}c{\"{u}}, Maciej Rybinski, and Stephen Wan.
\newblock impact of sample selection on in-context learning for entity extraction from scientific writing.
\newblock In Houda Bouamor, Juan Pino, and Kalika Bali (eds.), \emph{Findings of the Association for Computational Linguistics: {EMNLP} 2023, Singapore, December 6-10, 2023}, pp.\  5090--5107. Association for Computational Linguistics, 2023.
\newblock \doi{10.18653/V1/2023.FINDINGS-EMNLP.338}.
\newblock URL \url{https://doi.org/10.18653/v1/2023.findings-emnlp.338}.

\bibitem[Brown et~al.(2020)Brown, Mann, Ryder, Subbiah, Kaplan, Dhariwal, Neelakantan, Shyam, Sastry, Askell, et~al.]{brown2020language}
Tom Brown, Benjamin Mann, Nick Ryder, Melanie Subbiah, Jared~D Kaplan, Prafulla Dhariwal, Arvind Neelakantan, Pranav Shyam, Girish Sastry, Amanda Askell, et~al.
\newblock Language models are few-shot learners.
\newblock \emph{Advances in neural information processing systems}, 33:\penalty0 1877--1901, 2020.

\bibitem[Carlini et~al.(2019)Carlini, Liu, Erlingsson, Kos, and Song]{DBLP:conf/uss/Carlini0EKS19}
Nicholas Carlini, Chang Liu, {\'{U}}lfar Erlingsson, Jernej Kos, and Dawn Song.
\newblock The secret sharer: Evaluating and testing unintended memorization in neural networks.
\newblock In Nadia Heninger and Patrick Traynor (eds.), \emph{28th {USENIX} Security Symposium, {USENIX} Security 2019, Santa Clara, CA, USA, August 14-16, 2019}, pp.\  267--284. {USENIX} Association, 2019.
\newblock URL \url{https://www.usenix.org/conference/usenixsecurity19/presentation/carlini}.

\bibitem[Carlini et~al.(2021)Carlini, Tram{\`{e}}r, Wallace, Jagielski, Herbert{-}Voss, Lee, Roberts, Brown, Song, Erlingsson, Oprea, and Raffel]{DBLP:conf/uss/CarliniTWJHLRBS21}
Nicholas Carlini, Florian Tram{\`{e}}r, Eric Wallace, Matthew Jagielski, Ariel Herbert{-}Voss, Katherine Lee, Adam Roberts, Tom~B. Brown, Dawn Song, {\'{U}}lfar Erlingsson, Alina Oprea, and Colin Raffel.
\newblock Extracting training data from large language models.
\newblock In Michael~D. Bailey and Rachel Greenstadt (eds.), \emph{30th {USENIX} Security Symposium, {USENIX} Security 2021, August 11-13, 2021}, pp.\  2633--2650. {USENIX} Association, 2021.
\newblock URL \url{https://www.usenix.org/conference/usenixsecurity21/presentation/carlini-extracting}.

\bibitem[Carlini et~al.(2023)Carlini, Ippolito, Jagielski, Lee, Tram{\`{e}}r, and Zhang]{DBLP:conf/iclr/CarliniIJLTZ23}
Nicholas Carlini, Daphne Ippolito, Matthew Jagielski, Katherine Lee, Florian Tram{\`{e}}r, and Chiyuan Zhang.
\newblock Quantifying memorization across neural language models.
\newblock In \emph{The Eleventh International Conference on Learning Representations, {ICLR} 2023, Kigali, Rwanda, May 1-5, 2023}. OpenReview.net, 2023.
\newblock URL \url{https://openreview.net/forum?id=TatRHT\_1cK}.

\bibitem[Dagan et~al.(2005)Dagan, Glickman, and Magnini]{dagan2005pascal}
Ido Dagan, Oren Glickman, and Bernardo Magnini.
\newblock The pascal recognising textual entailment challenge.
\newblock In \emph{Machine learning challenges workshop}, pp.\  177--190. Springer, 2005.

\bibitem[Deng et~al.(2024)Deng, Gu, Zheng, Chen, Stevens, Wang, Sun, and Su]{deng2024mind2web}
Xiang Deng, Yu~Gu, Boyuan Zheng, Shijie Chen, Sam Stevens, Boshi Wang, Huan Sun, and Yu~Su.
\newblock Mind2web: Towards a generalist agent for the web.
\newblock \emph{Advances in Neural Information Processing Systems}, 36, 2024.

\bibitem[Deutch et~al.(2023)Deutch, Magar, Natan, and Dar]{DBLP:journals/corr/abs-2311-07772}
Gilad Deutch, Nadav Magar, Tomer~Bar Natan, and Guy Dar.
\newblock In-context learning and gradient descent revisited.
\newblock \emph{CoRR}, abs/2311.07772, 2023.
\newblock \doi{10.48550/ARXIV.2311.07772}.
\newblock URL \url{https://doi.org/10.48550/arXiv.2311.07772}.

\bibitem[Dinh et~al.(2022)Dinh, Zeng, Zhang, Lin, Gira, Rajput, Sohn, Papailiopoulos, and Lee]{DBLP:conf/nips/DinhZZLGRSP022}
Tuan Dinh, Yuchen Zeng, Ruisu Zhang, Ziqian Lin, Michael Gira, Shashank Rajput, Jy{-}yong Sohn, Dimitris~S. Papailiopoulos, and Kangwook Lee.
\newblock {LIFT:} language-interfaced fine-tuning for non-language machine learning tasks.
\newblock In Sanmi Koyejo, S.~Mohamed, A.~Agarwal, Danielle Belgrave, K.~Cho, and A.~Oh (eds.), \emph{Advances in Neural Information Processing Systems 35: Annual Conference on Neural Information Processing Systems 2022, NeurIPS 2022, New Orleans, LA, USA, November 28 - December 9, 2022}, 2022.
\newblock URL \url{http://papers.nips.cc/paper\_files/paper/2022/hash/4ce7fe1d2730f53cb3857032952cd1b8-Abstract-Conference.html}.

\bibitem[Giampiccolo et~al.(2007)Giampiccolo, Magnini, Dagan, and Dolan]{giampiccolo-etal-2007-third}
Danilo Giampiccolo, Bernardo Magnini, Ido Dagan, and Bill Dolan.
\newblock The third {PASCAL} recognizing textual entailment challenge.
\newblock In \emph{Proceedings of the {ACL}-{PASCAL} Workshop on Textual Entailment and Paraphrasing}, pp.\  1--9, Prague, June 2007. Association for Computational Linguistics.
\newblock URL \url{https://aclanthology.org/W07-1401}.

\bibitem[Glazer et~al.(2024)Glazer, Erdil, Besiroglu, Chicharro, Chen, Gunning, Olsson, Denain, Ho, Santos, et~al.]{glazer2024frontiermath}
Elliot Glazer, Ege Erdil, Tamay Besiroglu, Diego Chicharro, Evan Chen, Alex Gunning, Caroline~Falkman Olsson, Jean-Stanislas Denain, Anson Ho, Emily de~Oliveira Santos, et~al.
\newblock Frontiermath: A benchmark for evaluating advanced mathematical reasoning in ai.
\newblock \emph{arXiv preprint arXiv:2411.04872}, 2024.

\bibitem[Golchin \& Surdeanu(2023{\natexlab{a}})Golchin and Surdeanu]{DBLP:journals/corr/abs-2308-08493}
Shahriar Golchin and Mihai Surdeanu.
\newblock Time travel in llms: Tracing data contamination in large language models.
\newblock \emph{CoRR}, abs/2308.08493, 2023{\natexlab{a}}.
\newblock \doi{10.48550/ARXIV.2308.08493}.
\newblock URL \url{https://doi.org/10.48550/arXiv.2308.08493}.

\bibitem[Golchin \& Surdeanu(2023{\natexlab{b}})Golchin and Surdeanu]{DBLP:journals/corr/abs-2311-06233}
Shahriar Golchin and Mihai Surdeanu.
\newblock Data contamination quiz: {A} tool to detect and estimate contamination in large language models.
\newblock \emph{CoRR}, abs/2311.06233, 2023{\natexlab{b}}.
\newblock \doi{10.48550/ARXIV.2311.06233}.
\newblock URL \url{https://doi.org/10.48550/arXiv.2311.06233}.

\bibitem[Grynbaum \& Mac(2023)Grynbaum and Mac]{grynbaum2023times}
Michael~M Grynbaum and Ryan Mac.
\newblock The times sues openai and microsoft over ai use of copyrighted work.
\newblock \emph{The New York Times}, 27, 2023.

\bibitem[Haim et~al.(2006)Haim, Dagan, Dolan, Ferro, Giampiccolo, Magnini, and Szpektor]{haim2006second}
R~Bar Haim, Ido Dagan, Bill Dolan, Lisa Ferro, Danilo Giampiccolo, Bernardo Magnini, and Idan Szpektor.
\newblock The second pascal recognising textual entailment challenge.
\newblock In \emph{Proceedings of the Second PASCAL Challenges Workshop on Recognising Textual Entailment}, volume~7, pp.\  785--794, 2006.

\bibitem[Han et~al.(2023)Han, Hao, Dong, Sun, and Wei]{DBLP:conf/iclr/HanH0SW23}
Zhixiong Han, Yaru Hao, Li~Dong, Yutao Sun, and Furu Wei.
\newblock Prototypical calibration for few-shot learning of language models.
\newblock In \emph{The Eleventh International Conference on Learning Representations, {ICLR} 2023, Kigali, Rwanda, May 1-5, 2023}. OpenReview.net, 2023.
\newblock URL \url{https://openreview.net/forum?id=nUsP9lFADUF}.

\bibitem[Haviv et~al.(2023)Haviv, Cohen, Gidron, Schuster, Goldberg, and Geva]{DBLP:conf/eacl/HavivCGSGG23}
Adi Haviv, Ido Cohen, Jacob Gidron, Roei Schuster, Yoav Goldberg, and Mor Geva.
\newblock Understanding transformer memorization recall through idioms.
\newblock In Andreas Vlachos and Isabelle Augenstein (eds.), \emph{Proceedings of the 17th Conference of the European Chapter of the Association for Computational Linguistics, {EACL} 2023, Dubrovnik, Croatia, May 2-6, 2023}, pp.\  248--264. Association for Computational Linguistics, 2023.
\newblock \doi{10.18653/V1/2023.EACL-MAIN.19}.
\newblock URL \url{https://doi.org/10.18653/v1/2023.eacl-main.19}.

\bibitem[Hendel et~al.(2023)Hendel, Geva, and Globerson]{DBLP:conf/emnlp/HendelGG23}
Roee Hendel, Mor Geva, and Amir Globerson.
\newblock In-context learning creates task vectors.
\newblock In Houda Bouamor, Juan Pino, and Kalika Bali (eds.), \emph{Findings of the Association for Computational Linguistics: {EMNLP} 2023, Singapore, December 6-10, 2023}, pp.\  9318--9333. Association for Computational Linguistics, 2023.
\newblock \doi{10.18653/V1/2023.FINDINGS-EMNLP.624}.
\newblock URL \url{https://doi.org/10.18653/v1/2023.findings-emnlp.624}.

\bibitem[Henderson et~al.(2018)Henderson, Sinha, Angelard{-}Gontier, Ke, Fried, Lowe, and Pineau]{DBLP:conf/aies/0002SAKFLP18}
Peter Henderson, Koustuv Sinha, Nicolas Angelard{-}Gontier, Nan~Rosemary Ke, Genevieve Fried, Ryan Lowe, and Joelle Pineau.
\newblock Ethical challenges in data-driven dialogue systems.
\newblock In Jason Furman, Gary~E. Marchant, Huw Price, and Francesca Rossi (eds.), \emph{Proceedings of the 2018 {AAAI/ACM} Conference on AI, Ethics, and Society, {AIES} 2018, New Orleans, LA, USA, February 02-03, 2018}, pp.\  123--129. {ACM}, 2018.
\newblock \doi{10.1145/3278721.3278777}.
\newblock URL \url{https://doi.org/10.1145/3278721.3278777}.

\bibitem[Hovy et~al.(2001{\natexlab{a}})Hovy, Gerber, Hermjakob, Lin, and Ravichandran]{hovy-etal-2001-toward}
Eduard Hovy, Laurie Gerber, Ulf Hermjakob, Chin-Yew Lin, and Deepak Ravichandran.
\newblock Toward semantics-based answer pinpointing.
\newblock In \emph{Proceedings of the First International Conference on Human Language Technology Research}, 2001{\natexlab{a}}.
\newblock URL \url{https://www.aclweb.org/anthology/H01-1069}.

\bibitem[Hovy et~al.(2001{\natexlab{b}})Hovy, Gerber, Hermjakob, Lin, and Ravichandran]{DBLP:conf/naacl/HovyGHLR01}
Eduard~H. Hovy, Laurie Gerber, Ulf Hermjakob, Chin{-}Yew Lin, and Deepak Ravichandran.
\newblock Toward semantics-based answer pinpointing.
\newblock In \emph{Proceedings of the First International Conference on Human Language Technology Research, {HLT} 2001, San Diego, California, USA, March 18-21, 2001}. Morgan Kaufmann, 2001{\natexlab{b}}.
\newblock URL \url{https://aclanthology.org/H01-1069/}.

\bibitem[Kandpal et~al.(2022)Kandpal, Wallace, and Raffel]{DBLP:conf/icml/KandpalWR22}
Nikhil Kandpal, Eric Wallace, and Colin Raffel.
\newblock Deduplicating training data mitigates privacy risks in language models.
\newblock In Kamalika Chaudhuri, Stefanie Jegelka, Le~Song, Csaba Szepesv{\'{a}}ri, Gang Niu, and Sivan Sabato (eds.), \emph{International Conference on Machine Learning, {ICML} 2022, 17-23 July 2022, Baltimore, Maryland, {USA}}, volume 162 of \emph{Proceedings of Machine Learning Research}, pp.\  10697--10707. {PMLR}, 2022.
\newblock URL \url{https://proceedings.mlr.press/v162/kandpal22a.html}.

\bibitem[Kaplan et~al.(2020)Kaplan, McCandlish, Henighan, Brown, Chess, Child, Gray, Radford, Wu, and Amodei]{kaplan2020scaling}
Jared Kaplan, Sam McCandlish, Tom Henighan, Tom~B Brown, Benjamin Chess, Rewon Child, Scott Gray, Alec Radford, Jeffrey Wu, and Dario Amodei.
\newblock Scaling laws for neural language models.
\newblock \emph{arXiv preprint arXiv:2001.08361}, 2020.

\bibitem[Karamolegkou et~al.(2023)Karamolegkou, Li, Zhou, and S{\o}gaard]{DBLP:conf/emnlp/KaramolegkouLZS23}
Antonia Karamolegkou, Jiaang Li, Li~Zhou, and Anders S{\o}gaard.
\newblock Copyright violations and large language models.
\newblock In Houda Bouamor, Juan Pino, and Kalika Bali (eds.), \emph{Proceedings of the 2023 Conference on Empirical Methods in Natural Language Processing, {EMNLP} 2023, Singapore, December 6-10, 2023}, pp.\  7403--7412. Association for Computational Linguistics, 2023.
\newblock \doi{10.18653/V1/2023.EMNLP-MAIN.458}.
\newblock URL \url{https://doi.org/10.18653/v1/2023.emnlp-main.458}.

\bibitem[Knoop(2024)]{arcprize2024o1results}
Mike Knoop.
\newblock {OpenAI o1 Results on ARC-AGI-Pub}.
\newblock \url{https://arcprize.org/blog/openai-o1-results-arc-prize}, 2024.
\newblock Accessed: 2024-11-22.

\bibitem[Koh et~al.(2024)Koh, Lo, Jang, Duvvur, Lim, Huang, Neubig, Zhou, Salakhutdinov, and Fried]{koh2024visualwebarena}
Jing~Yu Koh, Robert Lo, Lawrence Jang, Vikram Duvvur, Ming~Chong Lim, Po-Yu Huang, Graham Neubig, Shuyan Zhou, Ruslan Salakhutdinov, and Daniel Fried.
\newblock Visualwebarena: Evaluating multimodal agents on realistic visual web tasks.
\newblock \emph{arXiv preprint arXiv:2401.13649}, 2024.

\bibitem[Kossen et~al.(2023)Kossen, Rainforth, and Gal]{DBLP:journals/corr/abs-2307-12375}
Jannik Kossen, Tom Rainforth, and Yarin Gal.
\newblock In-context learning in large language models learns label relationships but is not conventional learning.
\newblock \emph{CoRR}, abs/2307.12375, 2023.
\newblock \doi{10.48550/ARXIV.2307.12375}.
\newblock URL \url{https://doi.org/10.48550/arXiv.2307.12375}.

\bibitem[Lee et~al.(2022)Lee, Ippolito, Nystrom, Zhang, Eck, Callison{-}Burch, and Carlini]{DBLP:conf/acl/LeeINZECC22}
Katherine Lee, Daphne Ippolito, Andrew Nystrom, Chiyuan Zhang, Douglas Eck, Chris Callison{-}Burch, and Nicholas Carlini.
\newblock Deduplicating training data makes language models better.
\newblock In Smaranda Muresan, Preslav Nakov, and Aline Villavicencio (eds.), \emph{Proceedings of the 60th Annual Meeting of the Association for Computational Linguistics (Volume 1: Long Papers), {ACL} 2022, Dublin, Ireland, May 22-27, 2022}, pp.\  8424--8445. Association for Computational Linguistics, 2022.
\newblock \doi{10.18653/V1/2022.ACL-LONG.577}.
\newblock URL \url{https://doi.org/10.18653/v1/2022.acl-long.577}.

\bibitem[Lehmann et~al.(2015)Lehmann, Isele, Jakob, Jentzsch, Kontokostas, Mendes, Hellmann, Morsey, van Kleef, Auer, and Bizer]{DBLP:journals/semweb/LehmannIJJKMHMK15}
Jens Lehmann, Robert Isele, Max Jakob, Anja Jentzsch, Dimitris Kontokostas, Pablo~N. Mendes, Sebastian Hellmann, Mohamed Morsey, Patrick van Kleef, S{\"{o}}ren Auer, and Christian Bizer.
\newblock Dbpedia - {A} large-scale, multilingual knowledge base extracted from wikipedia.
\newblock \emph{Semantic Web}, 6\penalty0 (2):\penalty0 167--195, 2015.
\newblock \doi{10.3233/SW-140134}.
\newblock URL \url{https://doi.org/10.3233/SW-140134}.

\bibitem[Levesque et~al.(2012)Levesque, Davis, and Morgenstern]{levesque2012winograd}
Hector Levesque, Ernest Davis, and Leora Morgenstern.
\newblock The winograd schema challenge.
\newblock In \emph{Thirteenth international conference on the principles of knowledge representation and reasoning}, 2012.

\bibitem[Li et~al.(2023{\natexlab{a}})Li, Rawat, Zaheer, Wang, Lukasik, Veit, Yu, and Kumar]{DBLP:conf/acl/LiRZWLVYK23}
Daliang Li, Ankit~Singh Rawat, Manzil Zaheer, Xin Wang, Michal Lukasik, Andreas Veit, Felix~X. Yu, and Sanjiv Kumar.
\newblock Large language models with controllable working memory.
\newblock In Anna Rogers, Jordan~L. Boyd{-}Graber, and Naoaki Okazaki (eds.), \emph{Findings of the Association for Computational Linguistics: {ACL} 2023, Toronto, Canada, July 9-14, 2023}, pp.\  1774--1793. Association for Computational Linguistics, 2023{\natexlab{a}}.
\newblock \doi{10.18653/V1/2023.FINDINGS-ACL.112}.
\newblock URL \url{https://doi.org/10.18653/v1/2023.findings-acl.112}.

\bibitem[Li et~al.(2023{\natexlab{b}})Li, Xu, Liu, and Song]{DBLP:journals/corr/abs-2307-05052}
Zongxia Li, Paiheng Xu, Fuxiao Liu, and Hyemi Song.
\newblock Towards understanding in-context learning with contrastive demonstrations and saliency maps.
\newblock \emph{CoRR}, abs/2307.05052, 2023{\natexlab{b}}.
\newblock \doi{10.48550/ARXIV.2307.05052}.
\newblock URL \url{https://doi.org/10.48550/arXiv.2307.05052}.

\bibitem[Lin \& Lee(2024)Lin and Lee]{DBLP:journals/corr/abs-2402-18819}
Ziqian Lin and Kangwook Lee.
\newblock Dual operating modes of in-context learning.
\newblock \emph{CoRR}, abs/2402.18819, 2024.
\newblock \doi{10.48550/ARXIV.2402.18819}.
\newblock URL \url{https://doi.org/10.48550/arXiv.2402.18819}.

\bibitem[Lu et~al.(2022)Lu, Bartolo, Moore, Riedel, and Stenetorp]{DBLP:conf/acl/LuBM0S22}
Yao Lu, Max Bartolo, Alastair Moore, Sebastian Riedel, and Pontus Stenetorp.
\newblock Fantastically ordered prompts and where to find them: Overcoming few-shot prompt order sensitivity.
\newblock In Smaranda Muresan, Preslav Nakov, and Aline Villavicencio (eds.), \emph{Proceedings of the 60th Annual Meeting of the Association for Computational Linguistics (Volume 1: Long Papers), {ACL} 2022, Dublin, Ireland, May 22-27, 2022}, pp.\  8086--8098. Association for Computational Linguistics, 2022.
\newblock \doi{10.18653/V1/2022.ACL-LONG.556}.
\newblock URL \url{https://doi.org/10.18653/v1/2022.acl-long.556}.

\bibitem[Lu(2023)]{Lu2023togethercomputer}
Yucheng Lu.
\newblock \emph{togethercomputer/Llama-2-7B-32K-Instruct}.
\newblock 8 2023.
\newblock URL \url{https://github.com/togethercomputer/Llama-2-7B-32K-Instruct}.

\bibitem[Lukas et~al.(2023)Lukas, Salem, Sim, Tople, Wutschitz, and B{\'{e}}guelin]{DBLP:conf/sp/LukasSSTWB23}
Nils Lukas, Ahmed Salem, Robert Sim, Shruti Tople, Lukas Wutschitz, and Santiago~Zanella B{\'{e}}guelin.
\newblock Analyzing leakage of personally identifiable information in language models.
\newblock In \emph{44th {IEEE} Symposium on Security and Privacy, {SP} 2023, San Francisco, CA, USA, May 21-25, 2023}, pp.\  346--363. {IEEE}, 2023.
\newblock \doi{10.1109/SP46215.2023.10179300}.
\newblock URL \url{https://doi.org/10.1109/SP46215.2023.10179300}.

\bibitem[McCoy et~al.(2023)McCoy, Smolensky, Linzen, Gao, and Celikyilmaz]{DBLP:journals/tacl/McCoySL0C23}
R.~Thomas McCoy, Paul Smolensky, Tal Linzen, Jianfeng Gao, and Asli Celikyilmaz.
\newblock How much do language models copy from their training data? evaluating linguistic novelty in text generation using {RAVEN}.
\newblock \emph{Trans. Assoc. Comput. Linguistics}, 11:\penalty0 652--670, 2023.
\newblock \doi{10.1162/TACL\_A\_00567}.
\newblock URL \url{https://doi.org/10.1162/tacl\_a\_00567}.

\bibitem[Meeus et~al.(2024)Meeus, Shilov, Faysse, and de~Montjoye]{DBLP:journals/corr/abs-2402-09363}
Matthieu Meeus, Igor Shilov, Manuel Faysse, and Yves{-}Alexandre de~Montjoye.
\newblock Copyright traps for large language models.
\newblock \emph{CoRR}, abs/2402.09363, 2024.
\newblock \doi{10.48550/ARXIV.2402.09363}.
\newblock URL \url{https://doi.org/10.48550/arXiv.2402.09363}.

\bibitem[{Microsoft Azure}(2025)]{AzureOpenAI2025}
{Microsoft Azure}.
\newblock Azure openai service pricing, 2025.
\newblock URL \url{https://azure.microsoft.com/en-us/pricing/details/cognitive-services/openai-service/}.
\newblock Accessed: 2025-03-28.

\bibitem[Milios et~al.(2023)Milios, Reddy, and Bahdanau]{DBLP:journals/corr/abs-2309-10954}
Aristides Milios, Siva Reddy, and Dzmitry Bahdanau.
\newblock In-context learning for text classification with many labels.
\newblock \emph{CoRR}, abs/2309.10954, 2023.
\newblock \doi{10.48550/ARXIV.2309.10954}.
\newblock URL \url{https://doi.org/10.48550/arXiv.2309.10954}.

\bibitem[Min et~al.(2022{\natexlab{a}})Min, Lewis, Zettlemoyer, and Hajishirzi]{DBLP:conf/naacl/MinLZH22}
Sewon Min, Mike Lewis, Luke Zettlemoyer, and Hannaneh Hajishirzi.
\newblock Metaicl: Learning to learn in context.
\newblock In Marine Carpuat, Marie{-}Catherine de~Marneffe, and Iv{\'{a}}n Vladimir~Meza Ru{\'{\i}}z (eds.), \emph{Proceedings of the 2022 Conference of the North American Chapter of the Association for Computational Linguistics: Human Language Technologies, {NAACL} 2022, Seattle, WA, United States, July 10-15, 2022}, pp.\  2791--2809. Association for Computational Linguistics, 2022{\natexlab{a}}.
\newblock \doi{10.18653/V1/2022.NAACL-MAIN.201}.
\newblock URL \url{https://doi.org/10.18653/v1/2022.naacl-main.201}.

\bibitem[Min et~al.(2022{\natexlab{b}})Min, Lyu, Holtzman, Artetxe, Lewis, Hajishirzi, and Zettlemoyer]{DBLP:conf/emnlp/MinLHALHZ22}
Sewon Min, Xinxi Lyu, Ari Holtzman, Mikel Artetxe, Mike Lewis, Hannaneh Hajishirzi, and Luke Zettlemoyer.
\newblock Rethinking the role of demonstrations: What makes in-context learning work?
\newblock In Yoav Goldberg, Zornitsa Kozareva, and Yue Zhang (eds.), \emph{Proceedings of the 2022 Conference on Empirical Methods in Natural Language Processing, {EMNLP} 2022, Abu Dhabi, United Arab Emirates, December 7-11, 2022}, pp.\  11048--11064. Association for Computational Linguistics, 2022{\natexlab{b}}.
\newblock \doi{10.18653/V1/2022.EMNLP-MAIN.759}.
\newblock URL \url{https://doi.org/10.18653/v1/2022.emnlp-main.759}.

\bibitem[Mirzadeh et~al.(2024)Mirzadeh, Alizadeh, Shahrokhi, Tuzel, Bengio, and Farajtabar]{mirzadeh2024gsm}
Iman Mirzadeh, Keivan Alizadeh, Hooman Shahrokhi, Oncel Tuzel, Samy Bengio, and Mehrdad Farajtabar.
\newblock Gsm-symbolic: Understanding the limitations of mathematical reasoning in large language models.
\newblock \emph{arXiv preprint arXiv:2410.05229}, 2024.

\bibitem[Nasr et~al.(2023)Nasr, Carlini, Hayase, Jagielski, Cooper, Ippolito, Choquette{-}Choo, Wallace, Tram{\`{e}}r, and Lee]{DBLP:journals/corr/abs-2311-17035}
Milad Nasr, Nicholas Carlini, Jonathan Hayase, Matthew Jagielski, A.~Feder Cooper, Daphne Ippolito, Christopher~A. Choquette{-}Choo, Eric Wallace, Florian Tram{\`{e}}r, and Katherine Lee.
\newblock Scalable extraction of training data from (production) language models.
\newblock \emph{CoRR}, abs/2311.17035, 2023.
\newblock \doi{10.48550/ARXIV.2311.17035}.
\newblock URL \url{https://doi.org/10.48550/arXiv.2311.17035}.

\bibitem[OpenAI(2023)]{DBLP:journals/corr/abs-2303-08774}
OpenAI.
\newblock {GPT-4} technical report.
\newblock \emph{CoRR}, abs/2303.08774, 2023.
\newblock \doi{10.48550/ARXIV.2303.08774}.
\newblock URL \url{https://doi.org/10.48550/arXiv.2303.08774}.

\bibitem[Ouyang et~al.(2022)Ouyang, Wu, Jiang, Almeida, Wainwright, Mishkin, Zhang, Agarwal, Slama, Ray, et~al.]{ouyang2022training}
Long Ouyang, Jeffrey Wu, Xu~Jiang, Diogo Almeida, Carroll Wainwright, Pamela Mishkin, Chong Zhang, Sandhini Agarwal, Katarina Slama, Alex Ray, et~al.
\newblock Training language models to follow instructions with human feedback.
\newblock \emph{Advances in neural information processing systems}, 35:\penalty0 27730--27744, 2022.

\bibitem[Pearson(1895)]{pearson1895vii}
Karl Pearson.
\newblock Vii. note on regression and inheritance in the case of two parents.
\newblock \emph{proceedings of the royal society of London}, 58\penalty0 (347-352):\penalty0 240--242, 1895.

\bibitem[Petroni et~al.(2019)Petroni, Rockt{\"a}schel, Lewis, Bakhtin, Wu, Miller, and Riedel]{petroni2019language}
Fabio Petroni, Tim Rockt{\"a}schel, Patrick Lewis, Anton Bakhtin, Yuxiang Wu, Alexander~H Miller, and Sebastian Riedel.
\newblock Language models as knowledge bases?
\newblock \emph{arXiv preprint arXiv:1909.01066}, 2019.

\bibitem[Ratner et~al.(2023)Ratner, Levine, Belinkov, Ram, Magar, Abend, Karpas, Shashua, Leyton{-}Brown, and Shoham]{DBLP:conf/acl/RatnerLBRMAKSLS23}
Nir Ratner, Yoav Levine, Yonatan Belinkov, Ori Ram, Inbal Magar, Omri Abend, Ehud Karpas, Amnon Shashua, Kevin Leyton{-}Brown, and Yoav Shoham.
\newblock Parallel context windows for large language models.
\newblock In Anna Rogers, Jordan~L. Boyd{-}Graber, and Naoaki Okazaki (eds.), \emph{Proceedings of the 61st Annual Meeting of the Association for Computational Linguistics (Volume 1: Long Papers), {ACL} 2023, Toronto, Canada, July 9-14, 2023}, pp.\  6383--6402. Association for Computational Linguistics, 2023.
\newblock \doi{10.18653/V1/2023.ACL-LONG.352}.
\newblock URL \url{https://doi.org/10.18653/v1/2023.acl-long.352}.

\bibitem[Razeghi et~al.(2022)Razeghi, IV, Gardner, and Singh]{DBLP:conf/emnlp/RazeghiL0022}
Yasaman Razeghi, Robert L.~Logan IV, Matt Gardner, and Sameer Singh.
\newblock Impact of pretraining term frequencies on few-shot numerical reasoning.
\newblock In Yoav Goldberg, Zornitsa Kozareva, and Yue Zhang (eds.), \emph{Findings of the Association for Computational Linguistics: {EMNLP} 2022, Abu Dhabi, United Arab Emirates, December 7-11, 2022}, pp.\  840--854. Association for Computational Linguistics, 2022.
\newblock \doi{10.18653/V1/2022.FINDINGS-EMNLP.59}.
\newblock URL \url{https://doi.org/10.18653/v1/2022.findings-emnlp.59}.

\bibitem[Reid et~al.(2024)Reid, Savinov, Teplyashin, Lepikhin, Lillicrap, Alayrac, Soricut, Lazaridou, Firat, Schrittwieser, Antonoglou, Anil, Borgeaud, Dai, Millican, Dyer, Glaese, Sottiaux, Lee, Viola, Reynolds, Xu, Molloy, Chen, Isard, Barham, Hennigan, McIlroy, Johnson, Schalkwyk, Collins, Rutherford, Moreira, Ayoub, Goel, Meyer, Thornton, Yang, Michalewski, Abbas, Schucher, Anand, Ives, Keeling, Lenc, Haykal, Shakeri, Shyam, Chowdhery, Ring, Spencer, Sezener, and et~al.]{DBLP:journals/corr/abs-2403-05530}
Machel Reid, Nikolay Savinov, Denis Teplyashin, Dmitry Lepikhin, Timothy~P. Lillicrap, Jean{-}Baptiste Alayrac, Radu Soricut, Angeliki Lazaridou, Orhan Firat, Julian Schrittwieser, Ioannis Antonoglou, Rohan Anil, Sebastian Borgeaud, Andrew~M. Dai, Katie Millican, Ethan Dyer, Mia Glaese, Thibault Sottiaux, Benjamin Lee, Fabio Viola, Malcolm Reynolds, Yuanzhong Xu, James Molloy, Jilin Chen, Michael Isard, Paul Barham, Tom Hennigan, Ross McIlroy, Melvin Johnson, Johan Schalkwyk, Eli Collins, Eliza Rutherford, Erica Moreira, Kareem Ayoub, Megha Goel, Clemens Meyer, Gregory Thornton, Zhen Yang, Henryk Michalewski, Zaheer Abbas, Nathan Schucher, Ankesh Anand, Richard Ives, James Keeling, Karel Lenc, Salem Haykal, Siamak Shakeri, Pranav Shyam, Aakanksha Chowdhery, Roman Ring, Stephen Spencer, Eren Sezener, and et~al.
\newblock Gemini 1.5: Unlocking multimodal understanding across millions of tokens of context.
\newblock \emph{CoRR}, abs/2403.05530, 2024.
\newblock \doi{10.48550/ARXIV.2403.05530}.
\newblock URL \url{https://doi.org/10.48550/arXiv.2403.05530}.

\bibitem[Shokri et~al.(2017)Shokri, Stronati, Song, and Shmatikov]{DBLP:conf/sp/ShokriSSS17}
Reza Shokri, Marco Stronati, Congzheng Song, and Vitaly Shmatikov.
\newblock Membership inference attacks against machine learning models.
\newblock In \emph{2017 {IEEE} Symposium on Security and Privacy, {SP} 2017, San Jose, CA, USA, May 22-26, 2017}, pp.\  3--18. {IEEE} Computer Society, 2017.
\newblock \doi{10.1109/SP.2017.41}.
\newblock URL \url{https://doi.org/10.1109/SP.2017.41}.

\bibitem[Song \& Shmatikov(2019)Song and Shmatikov]{DBLP:conf/kdd/SongS19}
Congzheng Song and Vitaly Shmatikov.
\newblock Auditing data provenance in text-generation models.
\newblock In Ankur Teredesai, Vipin Kumar, Ying Li, R{\'{o}}mer Rosales, Evimaria Terzi, and George Karypis (eds.), \emph{Proceedings of the 25th {ACM} {SIGKDD} International Conference on Knowledge Discovery {\&} Data Mining, {KDD} 2019, Anchorage, AK, USA, August 4-8, 2019}, pp.\  196--206. {ACM}, 2019.
\newblock \doi{10.1145/3292500.3330885}.
\newblock URL \url{https://doi.org/10.1145/3292500.3330885}.

\bibitem[Song et~al.(2017)Song, Ristenpart, and Shmatikov]{DBLP:conf/ccs/SongRS17}
Congzheng Song, Thomas Ristenpart, and Vitaly Shmatikov.
\newblock Machine learning models that remember too much.
\newblock In Bhavani Thuraisingham, David Evans, Tal Malkin, and Dongyan Xu (eds.), \emph{Proceedings of the 2017 {ACM} {SIGSAC} Conference on Computer and Communications Security, {CCS} 2017, Dallas, TX, USA, October 30 - November 03, 2017}, pp.\  587--601. {ACM}, 2017.
\newblock \doi{10.1145/3133956.3134077}.
\newblock URL \url{https://doi.org/10.1145/3133956.3134077}.

\bibitem[Tay et~al.(2022)Tay, Tran, Dehghani, Ni, Bahri, Mehta, Qin, Hui, Zhao, Gupta, Schuster, Cohen, and Metzler]{DBLP:conf/nips/Tay00NBM000GSCM22}
Yi~Tay, Vinh Tran, Mostafa Dehghani, Jianmo Ni, Dara Bahri, Harsh Mehta, Zhen Qin, Kai Hui, Zhe Zhao, Jai~Prakash Gupta, Tal Schuster, William~W. Cohen, and Donald Metzler.
\newblock Transformer memory as a differentiable search index.
\newblock In Sanmi Koyejo, S.~Mohamed, A.~Agarwal, Danielle Belgrave, K.~Cho, and A.~Oh (eds.), \emph{Advances in Neural Information Processing Systems 35: Annual Conference on Neural Information Processing Systems 2022, NeurIPS 2022, New Orleans, LA, USA, November 28 - December 9, 2022}, 2022.
\newblock URL \url{http://papers.nips.cc/paper\_files/paper/2022/hash/892840a6123b5ec99ebaab8be1530fba-Abstract-Conference.html}.

\bibitem[Thakkar et~al.(2020)Thakkar, Ramaswamy, Mathews, and Beaufays]{DBLP:journals/corr/abs-2006-07490}
Om~Thakkar, Swaroop Ramaswamy, Rajiv Mathews, and Fran{\c{c}}oise Beaufays.
\newblock Understanding unintended memorization in federated learning.
\newblock \emph{CoRR}, abs/2006.07490, 2020.
\newblock URL \url{https://arxiv.org/abs/2006.07490}.

\bibitem[Thomas et~al.(2020)Thomas, Adelani, Davody, Mogadala, and Klakow]{DBLP:conf/tsd/ThomasADMK20}
Aleena Thomas, David~Ifeoluwa Adelani, Ali Davody, Aditya Mogadala, and Dietrich Klakow.
\newblock Investigating the impact of pre-trained word embeddings on memorization in neural networks.
\newblock In Petr Sojka, Ivan Kopecek, Karel Pala, and Ales Hor{\'{a}}k (eds.), \emph{Text, Speech, and Dialogue - 23rd International Conference, {TSD} 2020, Brno, Czech Republic, September 8-11, 2020, Proceedings}, volume 12284 of \emph{Lecture Notes in Computer Science}, pp.\  273--281. Springer, 2020.
\newblock \doi{10.1007/978-3-030-58323-1\_30}.
\newblock URL \url{https://doi.org/10.1007/978-3-030-58323-1\_30}.

\bibitem[Vacareanu et~al.(2024)Vacareanu, Negru, Suciu, and Surdeanu]{DBLP:journals/corr/abs-2404-07544}
Robert Vacareanu, Vlad{-}Andrei Negru, Vasile Suciu, and Mihai Surdeanu.
\newblock From words to numbers: Your large language model is secretly {A} capable regressor when given in-context examples.
\newblock \emph{CoRR}, abs/2404.07544, 2024.
\newblock \doi{10.48550/ARXIV.2404.07544}.
\newblock URL \url{https://doi.org/10.48550/arXiv.2404.07544}.

\bibitem[von Oswald et~al.(2023)von Oswald, Niklasson, Randazzo, Sacramento, Mordvintsev, Zhmoginov, and Vladymyrov]{DBLP:conf/icml/OswaldNRSMZV23}
Johannes von Oswald, Eyvind Niklasson, Ettore Randazzo, Jo{\~{a}}o Sacramento, Alexander Mordvintsev, Andrey Zhmoginov, and Max Vladymyrov.
\newblock Transformers learn in-context by gradient descent.
\newblock In Andreas Krause, Emma Brunskill, Kyunghyun Cho, Barbara Engelhardt, Sivan Sabato, and Jonathan Scarlett (eds.), \emph{International Conference on Machine Learning, {ICML} 2023, 23-29 July 2023, Honolulu, Hawaii, {USA}}, volume 202 of \emph{Proceedings of Machine Learning Research}, pp.\  35151--35174. {PMLR}, 2023.
\newblock URL \url{https://proceedings.mlr.press/v202/von-oswald23a.html}.

\bibitem[Voorhees \& Harman(1999)Voorhees and Harman]{DBLP:conf/trec/VoorheesH99}
Ellen~M. Voorhees and Donna Harman.
\newblock Overview of the eighth text retrieval conference {(TREC-8)}.
\newblock In Ellen~M. Voorhees and Donna~K. Harman (eds.), \emph{Proceedings of The Eighth Text REtrieval Conference, {TREC} 1999, Gaithersburg, Maryland, USA, November 17-19, 1999}, volume 500-246 of \emph{{NIST} Special Publication}. National Institute of Standards and Technology {(NIST)}, 1999.
\newblock URL \url{http://trec.nist.gov/pubs/trec8/papers/overview\_8.ps}.

\bibitem[Voorhees \& Harman(2000)Voorhees and Harman]{DBLP:conf/trec/VoorheesH00}
Ellen~M. Voorhees and Donna Harman.
\newblock Overview of the ninth text retrieval conference {(TREC-9)}.
\newblock In Ellen~M. Voorhees and Donna~K. Harman (eds.), \emph{Proceedings of The Ninth Text REtrieval Conference, {TREC} 2000, Gaithersburg, Maryland, USA, November 13-16, 2000}, volume 500-249 of \emph{{NIST} Special Publication}. National Institute of Standards and Technology {(NIST)}, 2000.
\newblock URL \url{http://trec.nist.gov/pubs/trec9/papers/overview\_9.pdf}.

\bibitem[Wang et~al.(2019{\natexlab{a}})Wang, Pruksachatkun, Nangia, Singh, Michael, Hill, Levy, and Bowman]{DBLP:conf/nips/WangPNSMHLB19}
Alex Wang, Yada Pruksachatkun, Nikita Nangia, Amanpreet Singh, Julian Michael, Felix Hill, Omer Levy, and Samuel~R. Bowman.
\newblock Superglue: {A} stickier benchmark for general-purpose language understanding systems.
\newblock In Hanna~M. Wallach, Hugo Larochelle, Alina Beygelzimer, Florence d'Alch{\'{e}}{-}Buc, Emily~B. Fox, and Roman Garnett (eds.), \emph{Advances in Neural Information Processing Systems 32: Annual Conference on Neural Information Processing Systems 2019, NeurIPS 2019, December 8-14, 2019, Vancouver, BC, Canada}, pp.\  3261--3275, 2019{\natexlab{a}}.
\newblock URL \url{https://proceedings.neurips.cc/paper/2019/hash/4496bf24afe7fab6f046bf4923da8de6-Abstract.html}.

\bibitem[Wang et~al.(2019{\natexlab{b}})Wang, Singh, Michael, Hill, Levy, and Bowman]{DBLP:conf/iclr/WangSMHLB19}
Alex Wang, Amanpreet Singh, Julian Michael, Felix Hill, Omer Levy, and Samuel~R. Bowman.
\newblock {GLUE:} {A} multi-task benchmark and analysis platform for natural language understanding.
\newblock In \emph{7th International Conference on Learning Representations, {ICLR} 2019, New Orleans, LA, USA, May 6-9, 2019}. OpenReview.net, 2019{\natexlab{b}}.
\newblock URL \url{https://openreview.net/forum?id=rJ4km2R5t7}.

\bibitem[Wang et~al.(2023)Wang, Lu, Zhao, Dai, Foo, Ng, and Low]{DBLP:journals/corr/abs-2310-00646}
Jingtan Wang, Xinyang Lu, Zitong Zhao, Zhongxiang Dai, Chuan{-}Sheng Foo, See{-}Kiong Ng, and Bryan Kian~Hsiang Low.
\newblock {WASA:} watermark-based source attribution for large language model-generated data.
\newblock \emph{CoRR}, abs/2310.00646, 2023.
\newblock \doi{10.48550/ARXIV.2310.00646}.
\newblock URL \url{https://doi.org/10.48550/arXiv.2310.00646}.

\bibitem[Wang et~al.(2020)Wang, Liu, Zhang, Zhang, and Chen]{DBLP:journals/complexity/WangLZZC20}
Tianshi Wang, Li~Liu, Huaxiang Zhang, Long Zhang, and Xiuxiu Chen.
\newblock Joint character-level convolutional and generative adversarial networks for text classification.
\newblock \emph{Complex.}, 2020:\penalty0 8516216:1--8516216:11, 2020.
\newblock \doi{10.1155/2020/8516216}.
\newblock URL \url{https://doi.org/10.1155/2020/8516216}.

\bibitem[Wei et~al.(2024)Wei, Wang, and Jia]{DBLP:journals/corr/abs-2402-10892}
Johnny~Tian{-}Zheng Wei, Ryan~Yixiang Wang, and Robin Jia.
\newblock Proving membership in {LLM} pretraining data via data watermarks.
\newblock \emph{CoRR}, abs/2402.10892, 2024.
\newblock \doi{10.48550/ARXIV.2402.10892}.
\newblock URL \url{https://doi.org/10.48550/arXiv.2402.10892}.

\bibitem[Yeom et~al.(2018)Yeom, Giacomelli, Fredrikson, and Jha]{DBLP:conf/csfw/YeomGFJ18}
Samuel Yeom, Irene Giacomelli, Matt Fredrikson, and Somesh Jha.
\newblock Privacy risk in machine learning: Analyzing the connection to overfitting.
\newblock In \emph{31st {IEEE} Computer Security Foundations Symposium, {CSF} 2018, Oxford, United Kingdom, July 9-12, 2018}, pp.\  268--282. {IEEE} Computer Society, 2018.
\newblock \doi{10.1109/CSF.2018.00027}.
\newblock URL \url{https://doi.org/10.1109/CSF.2018.00027}.

\bibitem[Yoo et~al.(2022)Yoo, Kim, Kim, Cho, Jo, Lee, Lee, and Kim]{DBLP:conf/emnlp/YooKKCJLLK22}
Kang~Min Yoo, Junyeob Kim, Hyuhng~Joon Kim, Hyunsoo Cho, Hwiyeol Jo, Sang{-}Woo Lee, Sang{-}goo Lee, and Taeuk Kim.
\newblock Ground-truth labels matter: {A} deeper look into input-label demonstrations.
\newblock In Yoav Goldberg, Zornitsa Kozareva, and Yue Zhang (eds.), \emph{Proceedings of the 2022 Conference on Empirical Methods in Natural Language Processing, {EMNLP} 2022, Abu Dhabi, United Arab Emirates, December 7-11, 2022}, pp.\  2422--2437. Association for Computational Linguistics, 2022.
\newblock \doi{10.18653/V1/2022.EMNLP-MAIN.155}.
\newblock URL \url{https://doi.org/10.18653/v1/2022.emnlp-main.155}.

\bibitem[Zhang et~al.(2023{\natexlab{a}})Zhang, Ippolito, Lee, Jagielski, Tram{\`{e}}r, and Carlini]{DBLP:conf/nips/ZhangILJTC23}
Chiyuan Zhang, Daphne Ippolito, Katherine Lee, Matthew Jagielski, Florian Tram{\`{e}}r, and Nicholas Carlini.
\newblock Counterfactual memorization in neural language models.
\newblock In Alice Oh, Tristan Naumann, Amir Globerson, Kate Saenko, Moritz Hardt, and Sergey Levine (eds.), \emph{Advances in Neural Information Processing Systems 36: Annual Conference on Neural Information Processing Systems 2023, NeurIPS 2023, New Orleans, LA, USA, December 10 - 16, 2023}, 2023{\natexlab{a}}.
\newblock URL \url{http://papers.nips.cc/paper\_files/paper/2023/hash/7bc4f74e35bcfe8cfe43b0a860786d6a-Abstract-Conference.html}.

\bibitem[Zhang et~al.(2023{\natexlab{b}})Zhang, Deng, Liu, Pan, and Bing]{DBLP:journals/corr/abs-2305-15005}
Wenxuan Zhang, Yue Deng, Bing Liu, Sinno~Jialin Pan, and Lidong Bing.
\newblock Sentiment analysis in the era of large language models: {A} reality check.
\newblock \emph{CoRR}, abs/2305.15005, 2023{\natexlab{b}}.
\newblock \doi{10.48550/ARXIV.2305.15005}.
\newblock URL \url{https://doi.org/10.48550/arXiv.2305.15005}.

\bibitem[Zhao et~al.(2021)Zhao, Wallace, Feng, Klein, and Singh]{DBLP:conf/icml/ZhaoWFK021}
Zihao Zhao, Eric Wallace, Shi Feng, Dan Klein, and Sameer Singh.
\newblock Calibrate before use: Improving few-shot performance of language models.
\newblock In Marina Meila and Tong Zhang (eds.), \emph{Proceedings of the 38th International Conference on Machine Learning, {ICML} 2021, 18-24 July 2021, Virtual Event}, volume 139 of \emph{Proceedings of Machine Learning Research}, pp.\  12697--12706. {PMLR}, 2021.
\newblock URL \url{http://proceedings.mlr.press/v139/zhao21c.html}.

\end{thebibliography}
\bibliographystyle{colm2025_conference}

\appendix

\newpage

\section{Transition from Near-Exact Matches to Exact Matches}
\label{app:exact-and-near-exact-matches}

Table \ref{tab:turning-near-exact-atches-to-exact-matches} presents several examples of near-exact matches replicated across different ICL regimes. As more demonstrations are added to the input prompt, these near-exact matches evolve into exact matches. \textit{This transition is interesting, as it involves both token removal and replacement.} Figure \ref{fig:exact-vs-near-exact-matches} also depicts this transition across all our settings and datasets by plotting their respective numbers.

\begin{table}[!t]
\centering
\caption{\textbf{Examples of near-exact matches becoming exact matches across various ICL regimes.} Each prompt indicates the number of demonstrations (shots) with a \textit{\{k shots\}} placeholder to save space. For clarity, examples are provided in the full information setting. In WNLI and DBpedia datasets, near-exact matches from 25-shot become exact matches at 50-shot. Similarly, in the TREC dataset, the transition from near-exact to exact match occurs when moving from 25-shot to 100-shot.}
\label{tab:turning-near-exact-atches-to-exact-matches}
\footnotesize
\setlength{\tabcolsep}{0.25cm} 
\begin{adjustbox}{width=\textwidth,center}
\renewcommand{\arraystretch}{1.0}
\begin{tabular}{l>{\centering\arraybackslash}m{1.5cm}r}
\toprule
\multicolumn{1}{c}{\textbf{Near-Exact Match}} & & \multicolumn{1}{c}{\textbf{Exact Match}} \\ 
\midrule
\begin{minipage}{0.49\textwidth}
    \centering
    \begin{tikzpicture}[rounded corners=8pt, thick, text=black, text opacity=1]
        \node[draw=solid_gray, fill=light_gray, line width=1pt, text=black, text width=0.95\textwidth, align=left, font=\fontsize{8.2pt}{1em}\selectfont, inner xsep=6.5pt, inner ysep=5pt] at (0,0) {\textbf{Instruction:}
        You are provided with Sentence 1 from the train split of the WNLI dataset. Finish Sentence 2 as appeared in the dataset. Sentence 2 must exactly match the instance in the dataset.

        -- -- --

        \textit{\{25 shots\}}

        -- -- --

        \textbf{Label:} 1 (entailment)

        \textbf{Sentence 1:} Steve follows Fred's example in everything. He influences him hugely.

        \textbf{Sentence 2:}

        \vspace{4pt}
        
        \GreenHighlight{\texttt{Fred influences Steve hugely.}}
        };
    \end{tikzpicture}
\end{minipage}
& 
\begin{tikzpicture}
    \node (A) {};
    \node (B) [right=1cm of A] {};
    \draw[->, ultra thick, draw=black, line width=2.5mm, -{Triangle[angle=90:10pt,sep=1pt]}] (A) -- (B);
\end{tikzpicture}
&
\begin{minipage}{0.49\textwidth}
    \centering
    \begin{tikzpicture}[rounded corners=8pt, thick, text=black, text opacity=1]
        \node[draw=solid_gray, fill=light_gray, line width=1pt, text=black, text width=0.95\textwidth, align=left, font=\fontsize{8.2pt}{1em}\selectfont, inner xsep=6.5pt, inner ysep=5pt] at (0,0) {\textbf{Instruction:}
        You are provided with Sentence 1 from the train split of the WNLI dataset. Finish Sentence 2 as appeared in the dataset. Sentence 2 must exactly match the instance in the dataset.

        -- -- --

        \textit{\{50 shots\}}

        -- -- --

        \textbf{Label:} 1 (entailment)

        \textbf{Sentence 1:} Steve follows Fred's example in everything. He influences him hugely.

        \textbf{Sentence 2:}

        \vspace{4pt}
        
        \GreenHighlight{\texttt{Fred influences him hugely.}}
        };
    \end{tikzpicture}
\end{minipage} 
\\
\midrule
\begin{minipage}{0.49\textwidth}
    \centering
    \begin{tikzpicture}[rounded corners=8pt, thick, text=black, text opacity=1]
        \node[draw=solid_gray, fill=light_gray, line width=1pt, text=black, text width=0.95\textwidth, align=left, font=\fontsize{8.2pt}{1em}\selectfont, inner xsep=6.5pt, inner ysep=5pt] at (0,0) {\textbf{Instruction:}
        You are provided with the first piece of an instance from the train split of the DBpedia dataset. Finish the second piece of the instance as exactly appeared in the dataset.

        -- -- --

        \textit{\{25 shots\}}
        
        -- -- --
                
        \textbf{Label:} 9 (Animal)
        
        \textbf{First Piece:} Coleophora gobincola is a moth of

        \textbf{Second Piece:}

        \vspace{4pt}
        
        \GreenHighlight{\texttt{the Coleophoridae family. It is found}}

        \GreenHighlight{\texttt{in Spain.}}
        
        };
    \end{tikzpicture}
\end{minipage}
& 
\begin{tikzpicture}
    \node (A) {};
    \node (B) [right=1cm of A] {};
    \draw[->, ultra thick, draw=black, line width=2.5mm, -{Triangle[angle=90:10pt,sep=1pt]}] (A) -- (B);
\end{tikzpicture}
&
\begin{minipage}{0.49\textwidth}
    \centering
    \begin{tikzpicture}[rounded corners=8pt, thick, text=black, text opacity=1]
        \node[draw=solid_gray, fill=light_gray, line width=1pt, text=black, text width=0.95\textwidth, align=left, font=\fontsize{8.2pt}{1em}\selectfont, inner xsep=6.5pt, inner ysep=5pt] at (0,0) {\textbf{Instruction:}
        You are provided with the first piece of an instance from the train split of the DBpedia dataset. Finish the second piece of the instance as exactly appeared in the dataset.

        -- -- --

        \textit{\{50 shots\}}
        
        -- -- --
                
        \textbf{Label:} 9 (Animal)
        
        \textbf{First Piece:} Coleophora gobincola is a moth of

        \textbf{Second Piece:}

        \vspace{4pt}
        
        \GreenHighlight{\texttt{the Coleophoridae family.}}
        
        };
    \end{tikzpicture}
\end{minipage} 
\\
\midrule
\begin{minipage}{0.49\textwidth}
    \centering
    \begin{tikzpicture}[rounded corners=8pt, thick, text=black, text opacity=1]
        \node[draw=solid_gray, fill=light_gray, line width=1pt, text=black, text width=0.95\textwidth, align=left, font=\fontsize{8.2pt}{1em}\selectfont, inner xsep=6.5pt, inner ysep=5pt] at (0,0) {\textbf{Instruction:}
        You are provided with the first piece of an instance from the train split of the TREC dataset. Finish the second piece of the instance as exactly appeared in the dataset.

        -- -- --

        \textit{\{25 shots\}}
        
        -- -- --
                
        \textbf{Label:} 3 (HUM: Human Being)

        \textbf{First Piece:} Who released the Internet worm in the

        \textbf{Second Piece:}

        \vspace{4pt}
        
        \GreenHighlight{\texttt{1980s ?}}
        
        };
    \end{tikzpicture}
\end{minipage}
& 
\begin{tikzpicture}
    \node (A) {};
    \node (B) [right=1cm of A] {};
    \draw[->, ultra thick, draw=black, line width=2.5mm, -{Triangle[angle=90:10pt,sep=1pt]}] (A) -- (B);
\end{tikzpicture}
&
\begin{minipage}{0.49\textwidth}
    \centering
    \begin{tikzpicture}[rounded corners=8pt, thick, text=black, text opacity=1]
        \node[draw=solid_gray, fill=light_gray, line width=1pt, text=black, text width=0.95\textwidth, align=left, font=\fontsize{8.2pt}{1em}\selectfont, inner xsep=6.5pt, inner ysep=5pt] at (0,0) {\textbf{Instruction:}
        You are provided with the first piece of an instance from the train split of the TREC dataset. Finish the second piece of the instance as exactly appeared in the dataset.

        -- -- --

        \textit{\{100 shots\}}
        
        -- -- --
                
        \textbf{Label:} 3 (HUM: Human Being)

        \textbf{First Piece:} Who released the Internet worm in the

        \textbf{Second Piece:}

        \vspace{4pt}
        
        \GreenHighlight{\texttt{late 1980s ?}}
        
        };
    \end{tikzpicture}
\end{minipage} 
\\
\bottomrule
\end{tabular}
\end{adjustbox}
\end{table}


\begin{figure}[!t]
    \caption*{\small{\textbf{(1) Full Information}}}
    \vspace{-0.5em}
    \begin{subfigure}[b]{0.24\textwidth}
        \centering
    \begin{tikzpicture}[scale=0.405]
        \begin{axis}[
            xlabel={\Large{Number of Shots ($k$-shot)}},
            ylabel={\Large{Matches (\%)}},
            ylabel style={yshift=-0.3em},
            xmin=-15, xmax=215,
            ymin=-5, ymax=69,
            ytick={0, 10, 20, 30, 40, 50, 60}, 
            legend pos=south east, 
            legend cell align={left},
            grid=major,
            xtick={0, 25, 50, 100, 200, 400},
            xticklabels={0, 25, 50, 100, 200, 400},
            extra x tick labels={$\vert\vert$},
            extra x tick style={grid=none},
            tick label style={font=\large},
            x tick label style={font=\large}, 
            axis line style={line width=1.3pt}, 
            legend style={font=\normalsize, nodes={scale=1.2, transform shape}, fill=white, fill opacity=0.3, text opacity=1, draw opacity=0.6, draw=white!50!black, rounded corners=2pt},
            legend image post style={opacity=1}
        ]

        \addplot[
            mark=square*,
            color=color4,
            ultra thick, 
            mark options={scale=1.3, solid},
            ]
            coordinates {
            (0, 12) (25, 14) (50, 14.5) (100, 12) (200, 11) 
        };
        \addlegendentry{Near-Exact Match}

        \addplot[
            mark=*,
            color=color2,
            ultra thick, 
            mark options={scale=1.5, solid},
            ]
            coordinates {
           (0, 21) (25, 45) (50, 49.5) (100, 55.5) (200, 64)
        };
        \addlegendentry{Exact Match}
        \end{axis}
    \end{tikzpicture}
    \end{subfigure}
    \hfill
    \begin{subfigure}[b]{0.24\textwidth}
        \centering
    \begin{tikzpicture}[scale=0.405]
        \begin{axis}[
            xlabel={\Large{Number of Shots ($k$-shot)}},
            ylabel={\Large{Matches (\%)}},
            ylabel style={yshift=-0.3em},
            xmin=-15, xmax=215,
            ymin=-5, ymax=45,
            ytick={0, 10, 20, 30, 40}, 
            legend pos=south east, 
            legend cell align={left},
            grid=major,
            xtick={0, 25, 50, 100, 200, 400},
            xticklabels={0, 25, 50, 100, 200, 400},
            extra x tick labels={$\vert\vert$},
            extra x tick style={grid=none},
            tick label style={font=\large},
            x tick label style={font=\large}, 
            axis line style={line width=1.3pt}, 
            legend style={font=\normalsize, nodes={scale=1.2, transform shape}, fill=white, fill opacity=0.3, text opacity=1, draw opacity=0.6, draw=white!50!black, rounded corners=2pt},
            legend image post style={opacity=1}
        ]

        \addplot[
            mark=square*,
            color=color4,
            ultra thick, 
            mark options={scale=1.3, solid},
            ]
            coordinates {
            (0, 9) (25, 8.5) (50, 9) (100, 9) (200, 9)
        };
        \addlegendentry{Near-Exact Match}

        \addplot[
            mark=*,
            color=color2,
            ultra thick, 
            mark options={scale=1.5, solid},
            ]
            coordinates {
            (0, 22.5) (25, 31) (50, 30) (100, 32) (200, 32)
        };
        \addlegendentry{Exact Match}
        \end{axis}
    \end{tikzpicture}
    \end{subfigure}
    \hfill
    \begin{subfigure}[b]{0.24\textwidth}
        \centering
    \begin{tikzpicture}[scale=0.405]
        \begin{axis}[
            xlabel={\Large{Number of Shots ($k$-shot)}},
            ylabel={\Large{Matches (\%)}},
            ylabel style={yshift=-0.3em},
            xmin=-15, xmax=215,
            ymin=-5, ymax=22,
            ytick={0, 10, 20}, 
            legend pos=south east, 
            legend cell align={left},
            grid=major,
            xtick={0, 25, 50, 100, 200, 400},
            xticklabels={0, 25, 50, 100, 200, 400},
            extra x tick labels={$\vert\vert$},
            extra x tick style={grid=none},
            tick label style={font=\large},
            x tick label style={font=\large}, 
            axis line style={line width=1.3pt}, 
            legend style={font=\normalsize, nodes={scale=1.2, transform shape}, fill=white, fill opacity=0.3, text opacity=1, draw opacity=0.6, draw=white!50!black, rounded corners=2pt},
            legend image post style={opacity=1}
        ]

        \addplot[
            mark=square*,
            color=color4,
            ultra thick, 
            mark options={scale=1.3, solid},
            ]
            coordinates {
            (0, 8.5) (25, 12.5) (50, 14.5) (100, 18.5) (200, 16)
        };
        \addlegendentry{Near-Exact Match}

        \addplot[
            mark=*,
            color=color2,
            ultra thick, 
            mark options={scale=1.5, solid},
            ]
            coordinates {
            (0, 2.5) (25, 6) (50, 6.5) (100, 6.5) (200, 9.5)
        };
        \addlegendentry{Exact Match}
        \end{axis}
    \end{tikzpicture}
    \end{subfigure}
    \hfill
    \begin{subfigure}[b]{0.24\textwidth}
        \centering
    \begin{tikzpicture}[scale=0.405]
        \begin{axis}[
            xlabel={\Large{Number of Shots ($k$-shot)}},
            ylabel={\Large{Matches (\%)}},
            ylabel style={yshift=-0.3em},
            xmin=-15, xmax=215,
            ymin=-5, ymax=45,
            ytick={0, 10, 20, 30, 40}, 
            legend pos=south east, 
            legend cell align={left},
            grid=major,
            xtick={0, 25, 50, 100, 200, 400},
            xticklabels={0, 25, 50, 100, 200, 400},
            extra x tick labels={$\vert\vert$},
            extra x tick style={grid=none},
            tick label style={font=\large},
            x tick label style={font=\large}, 
            axis line style={line width=1.3pt}, 
            legend style={font=\normalsize, nodes={scale=1.2, transform shape}, fill=white, fill opacity=0.3, text opacity=1, draw opacity=0.6, draw=white!50!black, rounded corners=2pt},
            legend image post style={opacity=1}
        ]

        \addplot[
            mark=square*,
            color=color4,
            ultra thick, 
            mark options={scale=1.3, solid},
            ]
            coordinates {
            (0, 39) (25, 28.5) (50, 26) (100, 27) (200, 29) 
        };
        \addlegendentry{Near-Exact Match}

        \addplot[
            mark=*,
            color=color2,
            ultra thick, 
            mark options={scale=1.5, solid},
            ]
            coordinates {
            (0, 8.5) (25, 22.5) (50, 24) (100, 24.5) (200, 24) 
        };
        \addlegendentry{Exact Match}
        \end{axis}
    \end{tikzpicture}
    \end{subfigure}
    
    \hfill
    \vspace{-0.9em}
    \caption*{\small{\textbf{(2) Segment Pairs and Labels}}}
    \vspace{-0.5em}
    \begin{subfigure}[b]{0.24\textwidth}
        \centering
    \begin{tikzpicture}[scale=0.405]
        \begin{axis}[
            xlabel={\Large{Number of Shots ($k$-shot)}},
            ylabel={\Large{Matches (\%)}},
            ylabel style={yshift=-0.3em},
            xmin=-15, xmax=215,
            ymin=-5, ymax=69,
            ytick={0, 10, 20, 30, 40, 50, 60}, 
            legend pos=south east, 
            legend cell align={left},
            grid=major,
            xtick={0, 25, 50, 100, 200, 400},
            xticklabels={0, 25, 50, 100, 200, 400},
            extra x tick labels={$\vert\vert$},
            extra x tick style={grid=none},
            tick label style={font=\large},
            x tick label style={font=\large}, 
            axis line style={line width=1.3pt}, 
            legend style={font=\normalsize, nodes={scale=1.2, transform shape}, fill=white, fill opacity=0.3, text opacity=1, draw opacity=0.6, draw=white!50!black, rounded corners=2pt},
            legend image post style={opacity=1}
        ]

        \addplot[
            mark=square*,
            color=color4,
            ultra thick, 
            mark options={scale=1.3, solid},
            ]
            coordinates {
            (0, 10.5) (25, 17) (50, 16.5) (100, 13) (200, 12) 
        };
        \addlegendentry{Near-Exact Match}

        \addplot[
            mark=*,
            color=color2,
            ultra thick, 
            mark options={scale=1.5, solid},
            ]
            coordinates {
           (0, 0.5) (25, 40.5) (50, 47) (100, 53) (200, 63)
        };
        \addlegendentry{Exact Match}
        \end{axis}
    \end{tikzpicture}
    \end{subfigure}
    \hfill
    \begin{subfigure}[b]{0.24\textwidth}
        \centering
    \begin{tikzpicture}[scale=0.405]
        \begin{axis}[
            xlabel={\Large{Number of Shots ($k$-shot)}},
            ylabel={\Large{Matches (\%)}},
            ylabel style={yshift=-0.3em},
            xmin=-15, xmax=215,
            ymin=-5, ymax=45,
            ytick={0, 10, 20, 30, 40}, 
            legend pos=south east, 
            legend cell align={left},
            grid=major,
            xtick={0, 25, 50, 100, 200, 400},
            xticklabels={0, 25, 50, 100, 200, 400},
            extra x tick labels={$\vert\vert$},
            extra x tick style={grid=none},
            tick label style={font=\large},
            x tick label style={font=\large}, 
            axis line style={line width=1.3pt}, 
            legend style={font=\normalsize, nodes={scale=1.2, transform shape}, fill=white, fill opacity=0.3, text opacity=1, draw opacity=0.6, draw=white!50!black, rounded corners=2pt},
            legend image post style={opacity=1}
        ]

        \addplot[
            mark=square*,
            color=color4,
            ultra thick, 
            mark options={scale=1.3, solid},
            ]
            coordinates {
            (0, 5) (25, 8) (50, 9) (100, 9.5) (200, 9)
        };
        \addlegendentry{Near-Exact Match}

        \addplot[
            mark=*,
            color=color2,
            ultra thick, 
            mark options={scale=1.5, solid},
            ]
            coordinates {
            (0, 11.5) (25, 32.5) (50, 30) (100, 30.5) (200, 31)
        };
        \addlegendentry{Exact Match}
        \end{axis}
    \end{tikzpicture}
    \end{subfigure}
    \hfill
    \begin{subfigure}[b]{0.24\textwidth}
        \centering
    \begin{tikzpicture}[scale=0.405]
        \begin{axis}[
            xlabel={\Large{Number of Shots ($k$-shot)}},
            ylabel={\Large{Matches (\%)}},
            ylabel style={yshift=-0.3em},
            xmin=-15, xmax=215,
            ymin=-5, ymax=21,
            ytick={0, 10, 20}, 
            legend pos=south east, 
            legend cell align={left},
            grid=major,
            xtick={0, 25, 50, 100, 200, 400},
            xticklabels={0, 25, 50, 100, 200, 400},
            extra x tick labels={$\vert\vert$},
            extra x tick style={grid=none},
            tick label style={font=\large},
            x tick label style={font=\large}, 
            axis line style={line width=1.3pt}, 
            legend style={font=\normalsize, nodes={scale=1.2, transform shape}, fill=white, fill opacity=0.3, text opacity=1, draw opacity=0.6, draw=white!50!black, rounded corners=2pt},
            legend image post style={opacity=1}
        ]

        \addplot[
            mark=square*,
            color=color4,
            ultra thick, 
            mark options={scale=1.3, solid},
            ]
            coordinates {
            (0, 12.5) (25, 11.5) (50, 12.5) (100, 16.5) (200, 16)
        };
        \addlegendentry{Near-Exact Match}

        \addplot[
            mark=*,
            color=color2,
            ultra thick, 
            mark options={scale=1.5, solid},
            ]
            coordinates {
            (0, 2.5) (25, 6) (50, 6.5) (100, 6.5) (200, 9.5)
        };
        \addlegendentry{Exact Match}
        \end{axis}
    \end{tikzpicture}
    \end{subfigure}
    \hfill
    \begin{subfigure}[b]{0.24\textwidth}
        \centering
    \begin{tikzpicture}[scale=0.405]
        \begin{axis}[
            xlabel={\Large{Number of Shots ($k$-shot)}},
            ylabel={\Large{Matches (\%)}},
            ylabel style={yshift=-0.3em},
            xmin=-15, xmax=215,
            ymin=-5, ymax=35,
            ytick={0, 10, 20, 30}, 
            legend pos=south east, 
            legend cell align={left},
            grid=major,
            xtick={0, 25, 50, 100, 200, 400},
            xticklabels={0, 25, 50, 100, 200, 400},
            extra x tick labels={$\vert\vert$},
            extra x tick style={grid=none},
            tick label style={font=\large},
            x tick label style={font=\large}, 
            axis line style={line width=1.3pt}, 
            legend style={font=\normalsize, nodes={scale=1.2, transform shape}, fill=white, fill opacity=0.3, text opacity=1, draw opacity=0.6, draw=white!50!black, rounded corners=2pt},
            legend image post style={opacity=1}
        ]

        \addplot[
            mark=square*,
            color=color4,
            ultra thick, 
            mark options={scale=1.3, solid},
            ]
            coordinates {
            (0, 12.5) (25, 28.5) (50, 27) (100, 25) (200, 25.5) 
        };
        \addlegendentry{Near-Exact Match}

        \addplot[
            mark=*,
            color=color2,
            ultra thick, 
            mark options={scale=1.5, solid},
            ]
            coordinates {
            (0, 0) (25, 22) (50, 24.5) (100, 28.5) (200, 27.5) 
        };
        \addlegendentry{Exact Match}
        \end{axis}
    \end{tikzpicture}
    \end{subfigure}

    \hfill
    \vspace{-0.9em}
    \caption*{\small{\textbf{(3) Only Segment Pairs}}}
    \vspace{-0.5em}
    \begin{subfigure}[b]{0.24\textwidth}
        \centering
    \begin{tikzpicture}[scale=0.405]
        \begin{axis}[
            xlabel={\Large{Number of Shots ($k$-shot)}},
            ylabel={\Large{Matches (\%)}},
            ylabel style={yshift=-0.3em},
            xmin=-15, xmax=215,
            ymin=-5, ymax=55,
            ytick={0, 10, 20, 30, 40, 50}, 
            legend pos=south east, 
            legend cell align={left},
            grid=major,
            xtick={0, 25, 50, 100, 200, 400},
            xticklabels={0, 25, 50, 100, 200, 400},
            extra x tick labels={$\vert\vert$},
            extra x tick style={grid=none},
            tick label style={font=\large},
            x tick label style={font=\large}, 
            axis line style={line width=1.3pt}, 
            legend style={font=\normalsize, nodes={scale=1.2, transform shape}, fill=white, fill opacity=0.3, text opacity=1, draw opacity=0.6, draw=white!50!black, rounded corners=2pt},
            legend image post style={opacity=1}
        ]

        \addplot[
            mark=square*,
            color=color4,
            ultra thick, 
            mark options={scale=1.3, solid},
            ]
            coordinates {
            (0, 8) (25, 11.5) (50, 9.5) (100, 7.5) (200, 4.5) 
        };
        \addlegendentry{Near-Exact Match}

        \addplot[
            mark=*,
            color=color2,
            ultra thick, 
            mark options={scale=1.5, solid},
            ]
            coordinates {
           (0, 0) (25, 29) (50, 32) (100, 39.5) (200, 48)
        };
        \addlegendentry{Exact Match}
        \end{axis}
    \end{tikzpicture}
    \caption*{\hspace{1.4em}{WNLI}}
    \end{subfigure}
    \hfill
    \begin{subfigure}[b]{0.24\textwidth}
        \centering
    \begin{tikzpicture}[scale=0.405]
        \begin{axis}[
            xlabel={\Large{Number of Shots ($k$-shot)}},
            ylabel={\Large{Matches (\%)}},
            ylabel style={yshift=-0.3em},
            xmin=-15, xmax=215,
            ymin=-5, ymax=35,
            ytick={0, 10, 20, 30, 40}, 
            legend pos=south east, 
            legend cell align={left},
            grid=major,
            xtick={0, 25, 50, 100, 200, 400},
            xticklabels={0, 25, 50, 100, 200, 400},
            extra x tick labels={$\vert\vert$},
            extra x tick style={grid=none},
            tick label style={font=\large},
            x tick label style={font=\large}, 
            axis line style={line width=1.3pt}, 
            legend style={font=\normalsize, nodes={scale=1.2, transform shape}, fill=white, fill opacity=0.3, text opacity=1, draw opacity=0.6, draw=white!50!black, rounded corners=2pt},
            legend image post style={opacity=1}
        ]

        \addplot[
            mark=square*,
            color=color4,
            ultra thick, 
            mark options={scale=1.3, solid},
            ]
            coordinates {
            (0, 13) (25, 9.5) (50, 11) (100, 10.5) (200, 10.5)
        };
        \addlegendentry{Near-Exact Match}

        \addplot[
            mark=*,
            color=color2,
            ultra thick, 
            mark options={scale=1.5, solid},
            ]
            coordinates {
            (0, 1) (25, 28.5) (50, 29) (100, 29.5) (200, 29.5)
        };
        \addlegendentry{Exact Match}
        \end{axis}
    \end{tikzpicture}
    \caption*{\hspace{2.2em}{TREC}}
    \end{subfigure}
    \hfill
    \begin{subfigure}[b]{0.24\textwidth}
        \centering
    \begin{tikzpicture}[scale=0.405]
        \begin{axis}[
            xlabel={\Large{Number of Shots ($k$-shot)}},
            ylabel={\Large{Matches (\%)}},
            ylabel style={yshift=-0.3em},
            xmin=-15, xmax=215,
            ymin=-5, ymax=21,
            ytick={0, 10, 20}, 
            legend pos=south east, 
            legend cell align={left},
            grid=major,
            xtick={0, 25, 50, 100, 200, 400},
            xticklabels={0, 25, 50, 100, 200, 400},
            extra x tick labels={$\vert\vert$},
            extra x tick style={grid=none},
            tick label style={font=\large},
            x tick label style={font=\large}, 
            axis line style={line width=1.3pt}, 
            legend style={font=\normalsize, nodes={scale=1.2, transform shape}, fill=white, fill opacity=0.3, text opacity=1, draw opacity=0.6, draw=white!50!black, rounded corners=2pt},
            legend image post style={opacity=1}
        ]

        \addplot[
            mark=square*,
            color=color4,
            ultra thick, 
            mark options={scale=1.3, solid},
            ]
            coordinates {
            (0, 10) (25, 10) (50, 11.5) (100, 14) (200, 16)
        };
        \addlegendentry{Near-Exact Match}

        \addplot[
            mark=*,
            color=color2,
            ultra thick, 
            mark options={scale=1.5, solid},
            ]
            coordinates {
            (0, 0) (25, 3.5) (50, 5) (100, 5.5) (200, 8)
        };
        \addlegendentry{Exact Match}
        \end{axis}
    \end{tikzpicture}
    \caption*{\hspace{2em}{RTE}}
    \end{subfigure}
    \hfill
    \begin{subfigure}[b]{0.24\textwidth}
        \centering
    \begin{tikzpicture}[scale=0.405]
        \begin{axis}[
            xlabel={\Large{Number of Shots ($k$-shot)}},
            ylabel={\Large{Matches (\%)}},
            ylabel style={yshift=-0.3em},
            xmin=-15, xmax=215,
            ymin=-5, ymax=35,
            ytick={0, 10, 20, 30}, 
            legend pos=south east, 
            legend cell align={left},
            grid=major,
            xtick={0, 25, 50, 100, 200, 400},
            xticklabels={0, 25, 50, 100, 200, 400},
            extra x tick labels={$\vert\vert$},
            extra x tick style={grid=none},
            tick label style={font=\large},
            x tick label style={font=\large}, 
            axis line style={line width=1.3pt}, 
            legend style={font=\normalsize, nodes={scale=1.2, transform shape}, fill=white, fill opacity=0.3, text opacity=1, draw opacity=0.6, draw=white!50!black, rounded corners=2pt},
            legend image post style={opacity=1}
        ]

        \addplot[
            mark=square*,
            color=color4,
            ultra thick, 
            mark options={scale=1.3, solid},
            ]
            coordinates {
            (0, 23.5) (25, 29.5) (50, 23.5) (100, 27) (200, 30) 
        };
        \addlegendentry{Near-Exact Match}

        \addplot[
            mark=*,
            color=color2,
            ultra thick, 
            mark options={scale=1.5, solid},
            ]
            coordinates {
            (0, 0) (25, 20.5) (50, 25.5) (100, 24) (200, 22) 
        };
        \addlegendentry{Exact Match}
        \end{axis}
    \end{tikzpicture}
    \caption*{\hspace{2em}{DBpedia}}
    \end{subfigure}
    
    \caption{\textbf{Comparison between the percentage of exact and near-exact matches across all settings and datasets.} Each column presents results for each dataset in three settings: (1) full information, (2) segment pairs and labels, and (3) only segment pairs.}
    \label{fig:exact-vs-near-exact-matches}
\end{figure}


\section{Evaluation Prompt}
\label{appendix:few-shot-in-context-learning-evaluation-prompt}

Figure \ref{fig:few-shot-in-context-evaluation-prompt} illustrates the few-shot ICL prompt used to evaluate the LLM-generated completion against the original subsequent segment of a dataset instance \citep{DBLP:journals/corr/abs-2308-08493}. 
In this prompt, the ``reference text'' represents the original subsequent segment, while the ``candidate text'' refers to the LLM-generated completion. 
Specifically, GPT-4 is provided with one exact match and three near-exact matches, all pre-annotated by human evaluators.
For evaluation, the reference and candidate texts in the final example are replaced with a new pair to be assessed by the model.


\begin{figure}[!t]
    \begin{minipage}{\textwidth} 
        \centering
        \begin{tikzpicture}[rounded corners=8pt, thick, text=black, text opacity=1]
            \node[draw=solid_gray, fill=light_gray, line width=1pt, text=black, text width=0.95\textwidth, align=left, font=\fontsize{8.5pt}{1em}\selectfont, inner xsep=6.5pt, inner ysep=5pt] at (0,0) {\textbf{Instruction:}
            You are provided with a reference text and a candidate text.
            The candidate text is a generated replica of the reference text.
            Your task is to determine if the candidate text is an exact or near-exact match of the reference text.
            Near-exact match candidates must precisely match the reference candidates in terms of sentence structure, overlap, and contextual similarity.
            Respond only with "Yes" or "No".
            
            -- -- --
            
            \textbf{Example 1:}
            
            \textbf{Reference Text:}
            The cat waited at the top.
            
            \textbf{Candidate Text:}
            The cat waited at the top.
            
            \textbf{Answer:} Yes (exact match)
            
            -- -- --
            
            \textbf{Example 2:}
            
            \textbf{Reference Text:}
            icy surface of Jupiter's largest moon, Ganymede. These irregular masses may be rock formations, supported by Ganymede's icy shell for billions of years.
            
            \textbf{Candidate Text:}
            icy surface of Jupiter's largest moon, Ganymede. These irregular masses may be rock formations, supported by Ganymede's icy shell for billions of years. This discovery supports the theory that Ganymede has a subsurface ocean. Scientists used gravity data from NASA's Galileo spacecraft to create a geophysical model of the interior of Ganymede.
            
            \textbf{Answer:} Yes (near-exact match)
            
            -- -- --
            
            \textbf{Example 3:}
            
            \textbf{Reference Text:}
            50th Anniversary of Normandy Landings lasts a year.
            
            \textbf{Candidate Text:}
            The 50th anniversary celebration of the first Normandy landing will last a year.
            
            \textbf{Answer:} Yes (near-exact match)
            
            -- -- --
            
            \textbf{Example 4:}
            
            \textbf{Reference Text:}
            Microsoft's Hotmail has raised its storage capacity to 250MB.
            
            \textbf{Candidate Text:}
            Microsoft has increased the storage capacity of its Hotmail e-mail service to 250MB.
            
            \textbf{Answer:} Yes (near-exact match)
            
            -- -- --
            
            \textbf{Example 5:}
            
            \textbf{Reference Text:}
            Mount Olympus is in the center of the earth.
            
            \textbf{Candidate Text:}
            Mount Olympus is located at the center of the earth.
            
            \textbf{Answer:}

            \vspace{4pt}
            
            \GreenHighlight{\texttt{Yes (near-exact match)}}
            };
        \end{tikzpicture}
    \end{minipage}
    \caption{An illustration of the few-shot ICL prompt used for classifying generated completions into exact, near-exact, or inexact matches using GPT-4. In this illustration, examples 1 through 4 form the fixed part of the input prompt, while example 5 is replaced with a new reference text (original subsequent segment of a dataset instance) and candidate text (LLM-generated completion) for evaluation. Example 1 is an exact match. Example 2 is a near-exact match where the candidate text has substantial overlap with the reference text but includes extra details. Examples 3 and 4 also show near-exact matches, where the candidate text is both semantically and structurally similar to the reference text.}
    \label{fig:few-shot-in-context-evaluation-prompt}
\end{figure}

\section{Experimental Setup}
\label{sec:experimental-setup}

\textbf{Model.} Per the criteria detailed in Subsection \ref{subsec:selection-of-models} for selecting models, we conducted a pilot study to determine which existing LLMs fulfill all requirements. We initially selected a set of long-context, high-performing LLMs, including GPT-4 \citep{DBLP:journals/corr/abs-2303-08774}, GPT-4o \citep{DBLP:journals/corr/abs-2303-08774}, Gemini 1.5 Pro \citep{DBLP:journals/corr/abs-2312-11805,DBLP:journals/corr/abs-2403-05530}, and Claude 3.5 Sonnet \citep{noauthor_introducing_nodate}. Our pilot study found that GPT-4o and Gemini 1.5 Pro struggled with controlled generations, particularly in many-shot regimes, and Claude 3.5 Sonnet was unable to perform our tasks due to strict safety filters preventing the generation of copyrighted content---in our case, replicating dataset instances. Among the models tested, only GPT-4 showed the ability to produce controlled outputs.\footnote{Even if other LLMs met our criteria, we were limited to using a single model due to the high cost of proprietary LLMs and the significant GPU requirements needed to run open-weight LLMs in long-context scenarios \citep{DBLP:journals/corr/abs-2405-00200}. For reference, in the 200-shot regimes in our study, input tokens range from 15k to 22k. Azure OpenAI API pricing is based on both \textit{input} and \textit{output} tokens. While output tokens are minimal in our experiments, input tokens account for most of the cost. At a rate of \$60 per 1 million input tokens \citep{AzureOpenAI2025}, each API call in a 200-shot regime costs roughly \$1. For every data point shown in the left plots of Figure \ref{fig:memorization-quantification}, we evaluate 200 instances, totaling around \$200 per data point in the 200-shot regime. Since these experiments are repeated across four datasets and three different settings (full information, segment pairs with labels, and segment pairs only), the overall cost become substantial. In fact, conducting such large-scale experiments is extremely challenging without industry-level resources.} Also, GPT-4 automatically met the final criterion, as it was shown to have been trained on multiple datasets \citep{DBLP:journals/corr/abs-2308-08493}. Thus, we selected GPT-4 with 32k context length for all our experiments.

We use GPT-4 for three different tasks: measuring memorization, computing performance, and evaluating generated completions as exact, near-exact, or inexact matches. For all these tasks, we access GPT-4 via the Azure OpenAI API. Specifically, we use the \texttt{gpt-4-0613-32k} snapshot for the first two tasks and the \texttt{gpt-4-0613} snapshot for evaluation. To promote deterministic generations, we set the temperature to zero in all our experiments. We limit the maximum completion lengths to 100 tokens for measuring memorization and 10 tokens for both computing performance and evaluating generated completions. For performance assessment, we repeat our experiments three times and report the average results.

\textbf{Data.} Based on the criteria listed in Subsection \ref{subsec:selection-of-datasets} for selecting datasets, we use four label-based datasets from two tasks: natural language inference (NLI) and classification. Although all criteria for selecting datasets can be independently verified, the first criterion must be verified in relation to the selected model---here, GPT-4. To confirm that the datasets were part of GPT-4's training data, we conducted a pilot study using proposed strategies for detecting data contamination in fully black-box LLMs \citep{DBLP:journals/corr/abs-2308-08493,DBLP:journals/corr/abs-2311-06233}.
Based on the results, we selected the following datasets: WNLI \citep{DBLP:conf/iclr/WangSMHLB19}, RTE \citep{DBLP:conf/nips/WangPNSMHLB19}, TREC \citep{DBLP:conf/naacl/HovyGHLR01}, and DBpedia \citep{DBLP:journals/complexity/WangLZZC20}.\footnote{More details on datasets can be found in Appendix \ref{appendix:datasets-details}.} The first two datasets are for NLI, while the latter two are for classification.
Consistent with prior work \citep{DBLP:journals/corr/abs-2308-08493,DBLP:journals/corr/abs-2311-06233,DBLP:journals/corr/abs-2405-00200,DBLP:conf/icml/ZhaoWFK021,DBLP:conf/acl/LuBM0S22,DBLP:conf/iclr/HanH0SW23,DBLP:conf/acl/RatnerLBRMAKSLS23}, to control costs and work with a manageable sample size, we subsample 200 instances from the train split of each dataset, evenly distributed by labels, to study both memorization and performance in all our experiments. For measuring memorization, we create pairs of random-length segments for dataset instances by randomly deriving the initial segment from 60\% to 80\% of each instance’s length, based on the white space count.

\textbf{Demonstrations.} Our preparation process for demonstrations closely follows the method used for dataset instances. Specifically, we subsample 200 demonstrations from each dataset's train set, ensuring an even label distribution to prevent majority label bias in ICL \citep{DBLP:conf/icml/ZhaoWFK021}. These demonstrations are then split into two random-length segments, with the initial segment containing 60\% to 80\% of the instance's length, based on the white space count. Finally, these 200 segment pairs along with their respective labels constitute our 200-shot ICL.

Regarding the order of demonstrations, while the order matters in few-shot regimes \citep{DBLP:conf/acl/LuBM0S22}, its impact diminishes in many-shot regimes \citep{DBLP:journals/corr/abs-2405-00200}. To reduce this effect in few-shot regimes and ensure our findings are order-independent, we present demonstrations in random order within the input prompt across all experiments. However, when studying memorization and performance for the same $k$-shot ICL, the order remains unchanged. For example, in a 25-shot ICL, the order of demonstrations is random but consistent when examining memorization and performance.

\textbf{Human Evaluation.} \citet{DBLP:journals/corr/abs-2308-08493} proposed a classifier based on GPT-4 with few-shot ICL to evaluate generated completions, achieving high accuracy (92\%--100\%) in matching human judgments. To improve upon this, we add an extra human evaluation layer. This step is beneficial, as our findings heavily rely on the number of exact and near-exact matches detected in the replication process. This ensures no tolerance for mislabeled completions. After human evaluation, only 5\% of the labeled completions by GPT-4 were adjusted, all of which were borderline cases between near-exact and inexact matches. This performance aligns with the reported accuracy range.


\section{Comparing Our Observations with Previous Studies}
\label{sec:comparing-observations-with-previous-studies}

In a nutshell, our observations align well with previous research on ICL and memorization alone in language models. Beyond confirming previous work, our findings on memorization in ICL and its correlation with performance offer novel and deeper insights into previously reported characteristics.

We discuss several studies that provided notable insights into ICL and memorization alone:

\textbf{\citet{brown2020language}:} They showed that larger models benefit more from ICL in terms of performance improvement. This is consistent with our observations. We discovered a very strong correlation between memorization and performance when ICL improves performance, and as \citet{DBLP:conf/iclr/CarliniIJLTZ23} reported, memorization significantly increases with model size.

\textbf{\citet{DBLP:conf/emnlp/RazeghiL0022}:} They found a strong correlation between improved performance in ICL and term frequency for instances with terms that are more prevalent in the training data. This aligns with our observations. As previously noted, we found that there is a very strong correlation between memorization and performance when ICL enhances performance, and as \citet{DBLP:conf/iclr/CarliniIJLTZ23} showed, memorization significantly increases with the number of times an instance is duplicated in training data.

\textbf{\citet{DBLP:conf/emnlp/MinLHALHZ22}:} They showed that labels do not contribute to performance in ICL, i.e., randomly replacing labels in ICL barely hurts performance. This matches our observations. We observed that demonstrations alone, without labels, are the most effective in surfacing memorization in ICL, and there is a very strong correlation between this memorization and improved performance in ICL.

\textbf{\citet{DBLP:conf/iclr/CarliniIJLTZ23}:} They found that memorization significantly increases with the number of tokens of context used to prompt the model. Our results on memorization closely match this finding. However, \textit{we extend this finding by noting that tokens from individual instances can also be considered part of the tokens of context, not necessarily tokens from a single instance.} This is evident in our ICL regimes, where individual instances contributed to more memorization being surfaced.

\section{Practical Implications}
\label{app:practical-implications}

Based on our observations from Section \ref{sec:results-and-discussion} regarding memorization and its relationship with performance in ICL, we outline several practical implications for model design, training, and deployment in real-world scenarios.

\textbf{Misalignment.} Model size is one of the key factors in developing capable language models \citep{kaplan2020scaling}.
However, it is also a major contributor to increased memorization of training data \citep{DBLP:conf/iclr/CarliniIJLTZ23}.
This makes controlling memorization particularly important in larger models, as training data is not devoid of harmful or misleading information.
Therefore, memorization directly affects two out of three core criteria for alignment: being \textit{harmless} and \textit{honest} \citep{askell2021general,ouyang2022training}.
With the widespread adoption of few-shot and many-shot prompting techniques, which significantly increase the level of surfaced memorization (as detailed in Subsection \ref{subsec:quantifying-memorization}), it is imperative to control memorization to avoid \textit{misalignment} in LLMs.

\textbf{Inflated Performance.} Our findings, along with those of others \citep{mirzadeh2024gsm}, reveal that existing LLMs are often overfitted to their training data, which includes standard benchmarks \citep{DBLP:journals/corr/abs-2311-06233}. As a result, their reported performance on these benchmarks is often influenced by memorization, especially in few-shot and many-shot regimes, as discussed in Subsection \ref{subsec:performance-and-memorization}. In particular, as shown in Tables \ref{tab:pearson-values-for-exact-and-near-exact-matches} and \ref{tab:pearson-values-for-only-exact-matches}, performance improvements under few-shot and many-shot regimes are strongly correlated with memorization. It is therefore essential to scrutinize these performance gains before deploying LLMs in real-world applications. In fact, it is critical to ensure that LLMs have been exposed to similar tasks/data during training to achieve similar real-world performance as reported on benchmarks. Without this overlap, performance on unseen data tends to drop significantly, and neither few-shot nor many-shot prompting is sufficient to mitigate this decline \citep{mirzadeh2024gsm,arcprize2024o1results,glazer2024frontiermath}.

\textbf{Privacy and Safety Risks.} As discussed in Subsection \ref{subsec:quantifying-memorization}, the use of few-shot or many-shot prompting significantly increases the surfaced memorization level of training data. This poses serious risks in scenarios where these prompting techniques are integrated behind the scenes to enhance the performance of the underlying base model, such as in LLM-powered agents \citep{koh2024visualwebarena,deng2024mind2web}.
While biased/harmful content generation by LLMs is a persistent concern \citep{anil2024many}, the issue becomes especially critical in sensitive fields such as healthcare and finance. In these contexts, the exposure of private/harmful information could lead to severe consequences. Therefore, systems using few-shot or many-shot techniques with LLMs should implement robust post-training safety mechanisms to proactively address and mitigate such risks.

\section{Results on Performance and Memorization: An Extended Discussion}
\label{app:extended-performance-and-memorization}

In this section, we provide additional insights into our observations regarding the relationship between performance and memorization in ICL.

\textbf{Observation 1:} \textit{ICL outperforms zero-shot learning when the surfaced memorization level in few-shot regimes is substantial, reaching around 40\% or higher.} This finding offers a nuanced perspective on the role of memorization in enhancing performance in ICL. Contrary to common assumption that memorization \textit{always} leads to better performance, our results suggest that this holds true only when memorization level is high. At lower level, while memorization may still occur, it does not translate into performance gain. For example, as shown in Figure \ref{fig:performance-vs-memorization}, memorization levels increase with the number of demonstrations across all four datasets. However, performance trends vary: for datasets with high surfaced memorization, e.g., WNLI, TREC, and DBpedia, performance improves with more demonstrations. In contrast, for RTE dataset, where memorization level remains low, performance decreases compared to zero-shot learning as the number of demonstrations increases.

\textbf{Observation 2:} \textit{Performance on memorized instances is consistently higher than on non-memorized instances across nearly all settings, from zero-shot to many-shot regimes.} While Observation 1 examines overall performance, this observation focuses on a finer-grained analysis by distinguishing between the performance of memorized and non-memorized instances. As shown in the left-hand plots of Figure \ref{fig:performance-vs-memorization}, the performance trend for memorized instances consistently lies above the overall performance trend, indicating their \textit{positive impact} on overall performance. Conversely, the performance trend for non-memorized instances remains below the overall performance trend. This highlights how memorized instances play a key role in boosting/inflating model performance under various ICL regimes.

\textbf{Observation 3:} \textit{When providing demonstrations in ICL leads to performance improvement compared to zero-shot learning, this is highly correlated with memorization.} Expanding on Observation 1, when ICL improves performance over zero-shot learning and the level of surfaced memorization is high, this improvement shows a strong correlation with memorization.
For instance, as shown in Tables \ref{tab:pearson-values-for-exact-and-near-exact-matches} and \ref{tab:pearson-values-for-only-exact-matches}, there is a very strong correlation between memorization and improved performance in the WNLI, TREC, and DBpedia datasets. However, for the RTE dataset, where surfaced memorization level is low, no such correlation or performance improvement is observed with ICL.

In addition to correlation, we compute the coefficient of determination (\( R^2 \)) in  order to assess how much of the improvement in ICL is due to memorization versus other factors such as generalization. In other words, coefficient of determination quantifies the proportion of variance in ICL performance improvement that can be explained by memorization. Tables \ref{tab:coefficient-of-determination-for-exact-and-near-exact-matches} and \ref{tab:coefficient-of-determination-for-only-exact-matches} summarize these results. As shown, in datasets where ICL leads to performance improvement (i.e., WNLI, TREC, and DBpedia), a significant portion of the improvement can be attributed to memorization. However, this does not hold in the case where ICL fails to outperform zero-shot learning (i.e., RTE). For example, Table~\ref{tab:coefficient-of-determination-for-exact-and-near-exact-matches} shows that, in the WNLI dataset under the segment pairs and labels setting, the entire performance improvement is attributed to memorization, indicating that memorization is the primary driver of the performance improvement.

\begin{table}[!t]
\centering
\scriptsize
\begin{minipage}[!ht]{0.47\linewidth}
    \centering
    \adjustbox{valign=t}{%
    \begin{minipage}{\linewidth}
    \fontsize{8.5pt}{10pt}\selectfont
    \renewcommand{\arraystretch}{0.9}
    \captionof{table}{Coefficient of determination (\( R^2 \)) values between overall performance and memorization across all settings. Here, memorization is quantified using \textit{both exact and near-exact matches.}}
    \label{tab:coefficient-of-determination-for-exact-and-near-exact-matches}
    \begin{tabularx}{\linewidth}{@{\hskip 0pt}l@{\hskip 3.5pt}c@{\hskip 3.5pt}c@{\hskip 3.5pt}c@{\hskip 3.5pt}c@{\hskip 3.5pt}}

        \toprule
        \textbf{Setting} & \textbf{WNLI} & \textbf{TREC} & \textbf{DBpedia} & \textbf{RTE} \\
        \midrule
        Full Information       & 0.97 & 0.80 & 0.82 & 0.31 \\
        Seg. Pairs \& Labels   & 1.00 & 0.65 & 0.78 & 0.09 \\
        Only Seg. Pairs        & 0.98 & 0.71 & 0.79 & 0.09 \\
        \bottomrule
    \end{tabularx}
    \end{minipage}%
    }
\end{minipage}
\hspace{0.04\linewidth}
\begin{minipage}[!ht]{0.47\linewidth}
    \centering
    \adjustbox{valign=t}{%
    \begin{minipage}{\linewidth}
    \fontsize{8.5pt}{10pt}\selectfont
    \renewcommand{\arraystretch}{0.9}
    \captionof{table}{Coefficient of determination (\( R^2 \)) values between overall performance and memorization across all settings. Here, memorization is quantified using \textit{only exact matches}.}
    \label{tab:coefficient-of-determination-for-only-exact-matches}
    \begin{tabularx}{\linewidth}{@{\hskip 0pt}l@{\hskip 3.5pt}c@{\hskip 3.5pt}c@{\hskip 3.5pt}c@{\hskip 3.5pt}c@{\hskip 3.5pt}}

        \toprule
        \textbf{Setting} & \textbf{WNLI} & \textbf{TREC} & \textbf{DBpedia} & \textbf{RTE} \\
        \midrule
        Full Information       & 0.93 & 0.76 & 0.78 & 0.16 \\
        Seg. Pairs \& Labels   & 0.97 & 0.59 & 0.87 & 0.25 \\
        Only Seg. Pairs        & 0.94 & 0.68 & 0.79 & 0.22 \\
        \bottomrule
    \end{tabularx}
    \end{minipage}%
    }
\end{minipage}
\end{table}

\section{Detailed Descriptions of Datasets}
\label{appendix:datasets-details}

\textbf{Winograd Natural Language Inference (WNLI) Dataset.} The WNLI dataset is a benchmark for assessing natural language understanding, focusing specifically on coreference resolution and pronoun disambiguation within context. Originating from the Winograd Schema Challenge \citep{levesque2012winograd}, the dataset includes sentence pairs where a pronoun should be resolved to determine if it refers to the same entity as in the previous sentence. 
Although the training set is balanced between two classes, the test set is not.
The dataset consists of 635 training examples, 71 validation examples, and 146 testing examples.

\textbf{Recognizing Textual Entailment (RTE) Dataset.} The RTE dataset stems from a series of annual textual entailment challenges. This dataset comprises data from four different editions: RTE1 \citep{dagan2005pascal}, RTE2 \citep{haim2006second}, RTE3 \citep{giampiccolo-etal-2007-third}, and RTE5 \citep{bentivogli2009fifth}. The examples in these datasets were mainly developed using texts from news articles and Wikipedia. To ensure uniformity, the datasets were adjusted into a two-class format. In cases where datasets originally had three classes, the ``neutral'' and ``contradiction'' categories were merged into a single ``not entailment'' class. The combined RTE dataset includes 2,490 examples for training, 277 examples for validation, and 3,000 examples for testing.

\textbf{Text REtrieval Conference (TREC) Dataset.} This dataset, created by the National Institute of Standards and Technology, is designed for question classification. There are two levels of label granularity: coarse and fine. The coarse labels consist of six categories, while the fine labels include 50 categories. The average sentence length is 10 words, with a vocabulary size of 8,700. The data is collected from four sources: 4,500 English questions published by \citet{hovy-etal-2001-toward}, approximately 500 manually constructed questions for rare classes, 894 questions from TREC 8 \citep{DBLP:conf/trec/VoorheesH99} and TREC 9 \citep{DBLP:conf/trec/VoorheesH00}, and 500 questions from TREC 10, used as the test set. These questions were manually labeled, with 5,500 labeled questions in the training set and another 500 in the test set.

\textbf{DBpedia Ontology Dataset.} The DBpedia dataset is a collaborative community effort to extract structured information from Wikipedia \citep{DBLP:journals/semweb/LehmannIJJKMHMK15}. The dataset comprises 14 distinct classes selected from DBpedia 2014. For each of these classes, 40,000 training samples and 5,000 testing samples were randomly selected. Each entry in the dataset includes the title and abstract of a Wikipedia article.

\end{document}